\documentclass[11pt]{article}

\usepackage[T1]{fontenc}
\usepackage[utf8]{inputenc}
\usepackage{lmodern}
\usepackage[margin=1in]{geometry}
\usepackage{amsmath,amssymb}
\usepackage{graphicx}
\usepackage{caption}
\usepackage[multiple]{footmisc}
\usepackage[round,authoryear]{natbib}
\usepackage{microtype}
\usepackage{xcolor}
\usepackage[colorlinks=true,linkcolor=blue!50!black,citecolor=blue!50!black,urlcolor=blue!50!black]{hyperref}
\usepackage{url}

\title{An Operationalization of Opaque Serial Depth\thanks{An earlier version of this work appeared on the Redwood Research blog on September~10, 2026: \url{https://www.redwoodresearch.org/blog/an-operationalization-of-opaque-serial-depth}.}}

\author{
  Nathan Sheffield \qquad Alek Westover \qquad Lukas Finnveden \\
  Alexa Pan \qquad Julian Stastny \qquad Ryan Greenblatt\thanks{Work done while at Redwood Research. Now at METR.}
  \\[1em]
  Redwood Research
}

\date{}

\begin{document}

\maketitle

\begin{abstract}
Chain-of-thought (CoT) is currently a valuable tool for overseeing AI models, but some architectural shifts could significantly reduce CoT monitorability. We operationalize a measure that serves as a proxy for the amount of unverbalized serial cognition a model can perform, as a specific instantiation of the notion of \emph{opaque serial depth}: roughly, the length of the longest path through a model's computation that does not pass through an interpretable bottleneck. Our main contribution is a particular standard for what counts as such a bottleneck. We define \emph{natural-language-rooted} (NL-rooted) nodes as nodes that were initialized from a pre-training prior to produce human-understandable outputs, whose output space has not been expanded beyond that prior, and whose outputs the model has not been modified to interpret as a different type of data. By construction, the resulting measure, NLS depth, can be adjudicated before training from an architecture, training recipe, and deployment configuration alone, and is insensitive to ordinary post-training such as prompting, distillation, and RL. We find that, across open-weight models, NLS depth correlates strongly with layer count, with some exceptions, and has grown slowly over recent years. It would likely grow slowly under continued scaling of standard transformers, while latent-reasoning architectures such as full-bandwidth transformers could increase it by a large amount. We work through how NLS depth applies to a range of existing and proposed architectures, discuss alternative operationalizations, and give a worked example bounding the depth of Gemma3-1B. We argue that AI developers should track and report the NLS depth of their models as a complement to direct measurements of CoT monitorability.
\end{abstract}

\section{Introduction}\label{introduction}

Currently, chain-of-thought (CoT) is a valuable tool for overseeing AI models \citep{korbak2025chain}. However, some architectural shifts could significantly reduce CoT monitorability \citep{taylor2026loss}. In a separate document,\footnote{\url{https://www.redwoodresearch.org/blog/proposal-for-tracking-architecture-on-monitorability}} we have proposed that AI companies should transparently share information about the degree to which their architectures may allow for latent reasoning and communication. To assist with this proposal, this paper operationalizes a measure that serves as a proxy for the amount of unverbalized serial cognition a model can perform. Our measure is a specific instantiation of the notion of ``opaque serial depth'', originally defined in a recent paper from GDM \citep{brown-cohen2026quantifying}.

To measure the opaque serial depth of a computation, \citet{brown-cohen2026quantifying} propose measuring the longest path in the computational graph which doesn't pass through some form of ``interpretable bottleneck''. Centrally, if one considers CoT tokens as ``interpretable'' but transformer hidden states as ``non-interpretable'', then \textbf{the opaque serial depth of a standard transformer is proportional to the number of layers}.\footnote{This concept is also sometimes used implicitly in the AI industry. For instance, in response to concerns about Astra's architecture, OpenAI's chief scientist \href{https://x.com/merettm/status/2095023204993490967}{recently announced} ``the depth of the computation graph for our present frontier models, including Astra, is within a factor of two of GPT-4''.}

Our main contribution in this paper is a particular standard for what counts as an ``interpretable bottleneck''. Roughly speaking, we want to consider nodes to be ``interpretable bottlenecks'' if they output text (as opposed to latent states), and were initialized from a pre-training prior. To make this notion more precise, we define the concept of \textbf{natural-language-rooted nodes (NL-rooted nodes)} in a computation graph.\footnote{This is similar to one of the notions of interpretable bottleneck discussed by Brown-Cohen et al. \citep[see][Appendix~A.2]{brown-cohen2026quantifying}, but is more specific.} In order for a set of nodes in a model's computational graph to count as NL-rooted, we require:

\begin{itemize}
\item \textbf{Natural-language initialization:} The model has been initialized from a pre-training prior to produce ``natural-language'' outputs at those nodes. (We have a very permissive notion of natural-language, which includes objects like ``code'', ``jsons'' and ``image patches'', and allows for distillation from models with garbled CoTs.)
\item \textbf{No expansion of the output space:} The only outputs of those nodes are tokens whose meanings were directly initialized from that pre-training prior. (Ruling out, for example, abstract CoT \citep{ramji2026thinking}, COCONUT \citep{hao2024training}, full-bandwidth transformers \citep{wang2026full}, and token superpositions \citep{zhang2025soft}.)
\item \textbf{No modifying the model to interpret the tokens as a different type of data:} The model should not be directly modified to use these tokens in ways radically disconnected from the pre-training prior. (This clause should not preclude natural training methods. It is primarily intended to rule out technical loopholes in the above bullet point, such as turning a latent reasoning model \citep{wang2026full} into a model that superficially reasons in natural language via encoding and decoding its activations as natural language tokens.)\footnote{For clarity, we also add a fourth criterion: \textbf{No treating the token as continuous intermediate states that are backpropagated through}. This should already have been implied by the above criteria, but it's important enough to clarify separately: No gradients should have been backpropagated from a layer later in the computational graph, through the nodes themselves. (This does not cover SFT or RL, since in these cases, the gradient starts at the node rather than being backpropagated through it from later layers.)}
\end{itemize}

We elaborate on these criteria \hyperref[definition-of-natural-language-rooted-nodes]{below}. Importantly, \textbf{all normal forms of post-training can be used without preventing nodes from being NL-rooted,} including prompting, distillation, and RL. (Even with aggressive length penalties or CoT supervision, and even if these forms of post-training make the CoT illegible in practice.)

Given this definition of NL-rooted nodes, we call our corresponding notion of opaque serial depth \textbf{``Natural-Language-rooted node''-Separated depth (NLS depth).} Beyond defining this term, we also argue that it \hyperref[nls-depth-tracks-concerning-architecture-changes]{tracks concerning architectural changes}, discuss how to apply NLS depth to a number of example architectures (in \hyperref[appendix-a-applying-nls-depth-to-plausible-architectures]{Appendix A}), and discuss some alternative operationalizations of opaque serial depth (in \hyperref[appendix-b-sensible-alternative-operationalizations]{Appendix B}).

Tracking the NLS depth of AI systems is not a replacement for tracking monitorability or no-CoT capabilities; it is an important complement. NLS depth has some advantages not shared by these other measurements. In particular:

\begin{itemize}
\item NLS depth is a relatively crisp and easy-to-quantify notion,\footnote{Determining NLS depth for novel architectures may require some human judgement, but we believe it requires less judgement than more direct measures of monitorability or opaque capabilities.} whereas it's unclear how to quantify or measure monitorability (e.g., what tasks should this be measured on, what elicitation strategies should be used).
\item NLS depth can be adjudicated before training begins, just given the architecture, training recipe, and intended deployment-time configuration of an AI system. This makes it easier to use in decisions about what models to train.
\end{itemize}

To achieve these properties, NLS depth is insensitive to some post-training choices that could in fact lead models to have significantly more or less interpretable CoT. However, we think NLS depth is a good (albeit imperfect) proxy\footnote{We're excited about work to develop even better proxies.} for the amount of opaque serial cognition that models can perform, which we think is highly \hyperref[appendix-c-systems-with-high-opaque-serial-reasoning-capabilities-are-likely-less-monitorable]{relevant for monitorability}. NLS depth \hyperref[nls-depth-tracks-concerning-architecture-changes]{does a good job} of separating architectures in the current natural language CoT paradigm (even with heavy scaling) from especially concerning architectures like scaled-up versions of full-bandwidth transformers \citep{wang2026full} (which can recur on opaque state throughout a full context window). Overall, we think it would be highly valuable to track the NLS depth of AI systems.

\section{Definition of natural-language-rooted nodes}\label{definition-of-natural-language-rooted-nodes}

Before defining NLS depth we must define what ``\textbf{natural-language-rooted}'' (NL-rooted) nodes in a computational graph are---these will serve as the ``interpretable bottlenecks'' in our definition of opaque serial depth.

Brown-Cohen et al. suggest two different candidate directions for defining what counts as an interpretable information bottleneck. Our concept of NL-rooted nodes is similar to the second one (suggested in \citealp[Appendix~A.2]{brown-cohen2026quantifying}), and has a similar motivation:

\begin{quote}
Intuitively, the key reason that LLM chains of thought are interpretable is because we train LLMs to predict natural language, providing the LLM with a ``natural language prior''. As a result, the model's semantics for the tokens in the chain of thought should mimic the semantics that humans would ascribe to those tokens, because that is what would be useful for predicting text that was generated by humans.

Inspired by this basic argument, we propose two criteria that together suggest that an intermediate node should be considered interpretable:

\begin{enumerate}
\item Natural language prior: The intermediate representation is directly optimized either (1) to imitate human-written text (e.g., via pre-training or supervised fine-tuning) or (2) to look good to humans (e.g., RLHF \citep{ziegler2019finetuning}).
\item Limited optimization towards other incentives: There are relatively weak training incentives other than towards the natural language prior (ideally none). Incentives that directly push against the natural language prior are worse than incentives that are orthogonal to it.
\end{enumerate}
\end{quote}

We pick a specific, fairly permissive version of the second criterion. Our intention is for this measure to track the difference between the current CoT paradigm and especially concerning architecture changes, rather than trying to draw a harder-to-define line about what type and degree of RL in the current paradigm should flip reasoning from counting as uninterpretable to interpretable.

In order for a set of nodes to be counted as NL-rooted, we require:

\begin{itemize}
\item \textbf{Natural-language initialization.}
\item \textbf{No expansion of the output space.}
\item \textbf{No modifying the model to interpret the tokens as a different type of data.}
\end{itemize}

We also stress a specific property that is implied by these but that's important enough to state separately: \textbf{Tokens in an NL rooted node should not be treated as a continuous intermediate state that is backpropagated through}.

We will now expand on each of these four properties.

\textbf{Natural-language initialization:} The model should have been initialized from a pre-training prior to imitate human-understandable media, with the ``NL-rooted'' nodes initially representing samples from this prior.

``Human-understandable media'' includes not only text, but also code, json files, images or audio (although note that multimodal CoT is somewhat more fraught; see discussion \hyperref[nls-depth-for-non-general-purpose-models]{below}). This admittedly makes ``natural-language-rooted'' a slightly unfortunate term.

``Human-understandable media'' is also presumptively assumed to include all outputs from NL-rooted nodes themselves, to allow for e.g. distillation from other models.\footnote{Though whether something counts as an NL rooted node for this purpose can be tricky to judge---see discussion towards the end of this section.} (Note that this means that the ``human-understandable media'' could, in some cases, end up being difficult to understand, if model CoTs grow to be very illegible.) If transparency about NLS depth is practiced, we would suggest that AI companies flag whether and how they use high-NLS-depth models in their distillation procedures, since it's possible that low-NLS-depth models could inherit some dangers when distilled from high-NLS-depth models.\footnote{In particular, there may be some serially-loaded capabilities that low-NLS-depth models wouldn't be able to do on their own, or learn when their outputs are being monitored by humans; but that they would be able to learn via distillation from high-NLS-depth models. An especially concerning skill like this would be learning how to use steganography in the model's CoT.}

\textbf{No expansion of the output space:} The nodes only represent tokens from an output space in which virtually\footnote{Our understanding is that some tokens sometimes fail to be used in pre-training for various technical reasons.} all tokens have been initialized from the pre-training prior, \emph{or} have individually determined human-understandable meanings, like \texttt{<EOS>}, \texttt{<UNK>}, \texttt{<think>}, \texttt{<PAD>}. This is intended to rule out techniques such as COCONUT \citep{hao2024training}, abstract CoT \citep{ramji2026thinking}, full-bandwidth transformers \citep{wang2026full}, and token superpositions \citep{zhang2025soft}. All of these techniques would be implemented by expanding the output space.\footnote{For COCONUT, to a continuous activation space. For abstract CoT, by increasing the vocabulary. For a full-bandwidth transformer, to a parallel continuous activation space (which could be done in another place in the computational graph, in which case the nodes passing natural language tokens may still be NL-rooted, but NLS depth would still be high via the other path in the computational graph). For token superpositions, the output would have to be expanded to either include the probability distribution over tokens (rather than the tokens themselves) or sums of embeddings (which would constitute embeddings).}

\textbf{No modifying the model to interpret the tokens as a different type of data:} Not only do we require that a ``natural-language-rooted node'' continue operating over the same output space, we specify that this output space still be interpreted roughly as natural language in the same format as the model used in pre-training (which would be ``continue a sequence of text'' for a normal transformer).

That is, a node doesn't count as natural-language-rooted if the model architecture has been changed such that one should expect the node's outputs to be interpreted as a completely different type of data than the human-understandable media it was trained to imitate. (Normal training methods like RL or prompt distillation might shift how the model uses its outputs, but aren't included in this definition because they won't predictably result in the tokens being used as a proxy encoding of some other underlying type of data.\footnote{Unless they're directly designed for this purpose. See below examples.} We are concerned here with methods that directly change the architecture, or train directly to cause the tokens to become a proxy for some other type of data.)

Here are some examples of cases this is meant to explicitly rule out:

\begin{enumerate}
\item Optimizing the outputs of an NL-rooted node to imitate (or otherwise very strongly optimize a model to encode information about) an AI model's opaque states. (Which we can operationalize as states that aren't NL-rooted.) This would include NLA (natural language autoencoders \citep{fraser-taliente2026natural}) that have been strongly optimized to optimally encode information about activations.
\item Taking multiple token samples, adding together the embeddings, and feeding the combination back into the model. This would count as changing the type of data since the sum of multiple different token embeddings is a type of input that the model has never encountered in pre-training, and it has no clear analogue in natural language.\footnote{And even if the sum happened to match an existing token embedding (or was discretized to do so) the semantics would be very different from the pre-training semantics. (The difference between ``X is the next word in this sequence'' and ``X is the equivalent of a combination of plausible words in this sequence in a particular embedding space''.)} Accordingly, the output would not count as NL-rooted.\footnote{Note that this means that the output can stop counting as NL-rooted due to how the output tokens are used later in the computation. More precisely, we can say that a bottleneck in a potentially-opaque path between two different computations may be considered opaque or non-opaque depending on how the information from that bottleneck gets used in the rest of the computation.}
\end{enumerate}

\textbf{No treating the token as continuous intermediate states that are backpropagated through to train model weights:} For current CoT models, each forward pass ends with a non-differentiable sampling step. As a result, when doing RL or SFT, gradients are never passed from the input layer of one pass through to the output layer of the previous pass.

We require the same property (no gradients from later nodes are backpropagated through the node to update model weights in training) in order for something to be counted as an NL-rooted node. We do not specify exactly how models sample tokens, but do require that the sampling operation itself similarly be treated as non-differentiable (algorithms that compute the loss directly from the token probabilities, such as policy gradients, are fine).

This criterion is probably superfluous given the above criteria.\footnote{In particular, the criterion of non-expanded output space prevents the expansion from discrete to continuous representations that would normally precede backpropagation. However, it's possible that some human-understandable media may start out being continuous from the start, e.g. images. Even so, backpropagation through the node should be ruled out by ``No modifying the model to interpret the tokens as a different type of data'' since, after local perturbations optimized for providing useful representations to the next layer, an output would represent something quite different from anything that was encountered in pre-training.} However, we still think this criterion is worth mentioning separately, since backpropagation through representations seems like an especially likely way that they could be made hard-to-monitor. The concern is that backpropagating through NL-rooted nodes would cause changes in the nodes' representations that wouldn't correspond to any human-understandable semantics. Much more so than with RL (or e.g., evolutionary algorithms), where the search primarily happens within a distribution descended from pre-training.

The same restriction would apply to other algorithms that aren't quite backpropagation, but that also treat the nodes as continuous and search for small local changes to them that optimize their usefulness for the layer in front of them. Also, backpropagating through a node that's \emph{shortly} before an NL-rooted node in the computational graph may cause it to no longer be NL-rooted. This should rarely matter for practical NLS depth calculations, though could matter for distillation. See the footnote for details.\footnote{Consider a situation where you have a set of nodes that sample from a distribution for predicting natural language, and you backpropagate through that distribution or the activations that produce it from later in the network (perhaps you pass the activations to the next layer like COCONUT, and backpropagate). Then this will predictably influence the sample even if the sampling nodes themselves are not backpropagated through. Accordingly, this should be sufficient to count a set of nodes as no longer being NL-rooted. This should generally not matter much for the purpose of determining NLS depth, since models like this would generally have high NLS depth \emph{via the path through its latents} even if it doesn't have high NLS depth \emph{through this token sample}. One situation where this matters is for whether distilling off of the full-bandwidth transformer \citep{wang2026full} model would count as distilling from NL-rooted nodes or not. However, this isn't extremely important to settle in advance because we can already get the monitorability-relevant information if AI companies disclose when they are performing distillation from a high-NLS-depth model. Nevertheless, we will note that this would be quite subtle to adjudicate. For example, the middle parts of a transformer are constantly backpropagated through to produce more useful representations for late-sequence passes through the model, but its output is still NL-rooted. Similarly, if a full-bandwidth transformer \citep{wang2026full} was pre-trained to imitate text while also having its latents backpropagated through in a recurrent fashion, its text outputs would still count as being NL-rooted. The difference with the COCONUT example is that the latents of the full bandwidth transformer would have been optimized concurrently with optimization that pressures the text outputs to become and remain a good representation of natural language. Accordingly, when judging what counts as an NL-rooted node, it can be important to consider whether the computation producing it was backpropagated through for other purposes concurrently with or after the pre-training initialization.}

For worked examples of how to adjudicate whether something counts as a natural-language-rooted node or not, see \hyperref[appendix-a-applying-nls-depth-to-plausible-architectures]{Appendix A}.

\section{Definition of NLS depth}\label{definition-of-nls-depth}

We now build upon the definition of natural-language-rooted nodes to define the main measure of interest, ``Natural-Language-rooted node''-Separated depth (NLS depth).

\textbf{NLS depth:} A computation has NLS depth at most $D$ if the function it implements can be represented by an \emph{admissible circuit} in which every \emph{opaque path} has length at most $D$.

Three key concepts invoked in this definition are:

\begin{itemize}
\item \textbf{Admissible circuit:} A circuit with gates that are arbitrary functions from two 32-bit values to a single 32-bit value, consisting of at most $10^{26}$ gates. (The choice of ``32'' bits in particular, and the choice of ``$10^{26}$'' gates in particular, could be replaced by other reasonable choices and mostly does not make a substantive difference to the definition.\footnote{We need to specify some type of gate, and pick pairs of 32-bit values to a single 32-bit value, but alternative choices of gates would mostly not make any substantial difference to the proposal (see \hyperref[appendix-g-issues-with-low-flop-serially-intense-computations]{Appendix G} for some discussion of one exception). Our approach here differs slightly from the Brown-Cohen et al.'s approach, see commentary right below the definition for more discussion.
There also needs to be some size limit (in our case $10^{26}$ gates), because there exists an exponentially-large low-depth circuit for any function. But complexity theorists believe there likely aren't good ways to trade modest amounts of width for better depth in general, so a very loose bound here is fine. A circuit of size $10^{26}$ would correspond to $10^{26}$ FP32 FLOP (for a single forward pass) or around $3 \cdot 10^{27}$ FP4 FLOP. For a full-context length forward pass, this could be 7 OOMs higher. (The bound is likely still 6 OOMs larger than is necessary for current frontier models. It could be adjusted upwards if needed. This may be necessary for a computational graph representing an agent swarm.) Note that our approach here differs from the GDM paper, which instead says that the total number of gates must be at most polynomial in the total number of parameters. We avoid this definition because in order to be meaningful, it requires specifying a \emph{family} of neural networks with increasingly large numbers of parameters, which adds unnecessary complexity.}) Note that a neural network (or system of neural networks) can easily be converted into an admissible circuit.\footnote{This circuit would take in the input of the neural network(s), and produce the output of the neural network(s)---we consider model weights as constants, so they can be hardwired into the gates as opposed to having to be passed as a separate input.}
\item \textbf{Opaque path:} Any path in a circuit (with edges directed from inputs towards outputs) which doesn't involve any circuit node marked as an interpretable bottleneck (i.e. a natural-language-rooted node).
\item \textbf{Can be represented by:} The strictest reading would require the circuit to compute exactly the same function as the neural network. We relax this slightly: the circuit may differ in ways that don't materially change the model's outputs, such as reordering reductions, altering floating-point rounding, or removing quantization. It may not differ in ways that materially change behavior. For instance a shallower model that merely matches the original's benchmark performance (e.g., via distillation) does not count as a representation of it.\footnote{To reduce variance in how this clause is interpreted, NLS depth estimates of models with secret architectures should ideally be reported as a \emph{multiple} of the NLS depth of models with publicly known architectures, with this clause interpreted in the same way for both calculations. Both NLS depth estimates should also be optimized to a similar degree and in a similar way. This reduces variance from the perspective of observers who don't know how much further optimized the bounds can be.}
\end{itemize}

This is very similar to the definition given by \citet{brown-cohen2026quantifying}; the only difference is in our choice of how to define an \emph{admissible circuit}.\footnote{Most importantly, we suggest that the gates should be arbitrary functions between fixed-precision inputs and outputs, whereas GDM's gate permissible set is: ``(1) Associative binary operations applied to two real numbers e.g. addition, multiplication, max, or min. (2) Piecewise functions of a single real number, where each piece is an analytic function.'' For the purposes of making our definition meaningful up to constant factors, it's important to fix a precision for the gates. There isn't a good reason to restrict to associative binary operations instead of arbitrary operations on two fixed-precision numbers. So, our choice of gate set makes the notion of opaque serial depth more meaningful up to constant factors, and simpler. One nice property of this definition is that it permits having arbitrary activation functions on a single number without increasing the depth. We also suggest a fixed but somewhat arbitrary $10^{26}$ ceiling on size rather than an unspecified polynomial size-limit.}

This definition can technically be applied to any computation, but in practice we want to use this definition to measure the NLS depth of \textbf{AI systems}. Intuitively, an ``AI system'' is some machine learning system that is deployed to solve a problem (e.g., a math problem). A central example of an AI system is an LLM run on a prompt for one context-length. Some other objects that count as AI systems include: an agent run with compaction, an agent that stores memories, an agent that performs many backpropagations over the course of an inference rollout (e.g., as in TTT \citep{tandon2025end}), and a swarm of communicating agents. An example of a computation that we wouldn't want to consider an AI system is ``the entire process of training an AI model''. When an LLM interacts with humans or with a computer via tool calls, we also do not consider that human or computer to be part of the AI system.

Here is some intuition about the NLS depth of some of the AI systems mentioned above:

\begin{itemize}
\item For a model that reasons in natural-language CoT, the NLS depth of an entire inference rollout is the same as the NLS depth of a single forward pass (done at maximum context length). This is true even if the inference rollout involves compacting the context to a textual summary of the context, storing textual memories during the context, and even if the inference rollout involves communicating in text with other agents.
\item For a model that does latent reasoning (like Meta's COCONUT \citep{hao2024training}), the NLS depth of an inference rollout is the number of latent reasoning steps times the NLS depth of a single forward pass. Similarly, for a looped transformer, the NLS depth is the depth of a loop iteration times the number of loops. This is true in each case even if tokens are generated in parallel with the latent state, since there is still some long path through opaque state.
\item If a model stores and later reads opaque memories, then the longest chain of opaque memories would determine the NLS depth.
\item Similarly, if you had a swarm of agents that were communicating using opaque state, then the longest chain of opaque state would determine the NLS depth.
\end{itemize}

The NLS depth of a given AI system can vary across different deployment configurations and inputs. Current LLMs have slightly higher depth at longer context lengths (the scaling of depth with sequence length is generally quite slow).\footnote{For a standard transformer architecture with dense attention, depth grows proportional to log(sequence length) in the limit. (This is related to how summing $n$ terms takes depth $\log n$ because you sum them in a tree.) However, for realistic configurations it's slower, as log(sequence length) is added to a term that scales with width, which dominates. A model with hidden dimension 4000 and that uses non-sparse attention would have a less than 2x factor difference in its opaque serial depth when evaluated at context length $1$ vs.\ $10^9$. In practice, models use sparse attention, much shorter context lengths, and higher width, so depth increasing with context length isn't important for transformers with standard architectures. However, some non-standard architectures do scale closer to log(sequence length) at realistic configurations.} More importantly, some architectures could allow for variable length forward passes \citep{geiping2025scaling}. We expect that the maximum NLS depth used in practice is the most important version to track (but the rough distribution of NLS depth as well as the maximum readily feasible and performant NLS depth could also be informative).

\begin{figure}[t]
  \centering
  \includegraphics[width=0.9\linewidth]{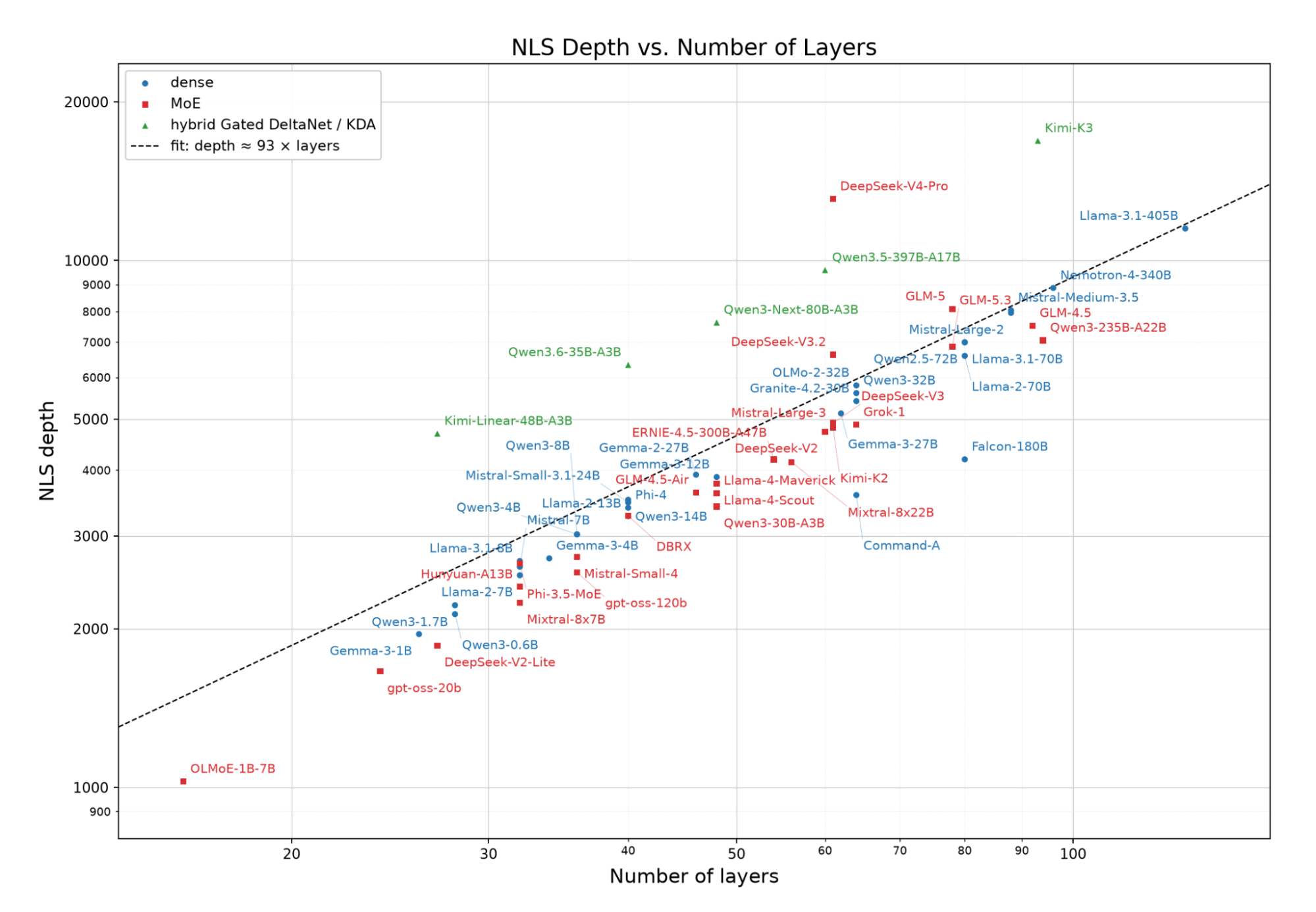}
  \caption[There's a strong correlation between NLS depth and the number of layers in open-source models.]{There's a strong correlation between NLS depth and the number of layers in open-source models.}
  \label{fig:1}
\end{figure}

NLS depth is a fairly natural property of a model, and it's fairly easy to compute a good bound on a model's depth---we give an example of this computation for Gemma3-1B in \hyperref[appendix-d-worked-example-of-bounding-depth]{Appendix D}. In particular, given this paper and the details of an architecture, existing consumer AI models are capable of computing bounds we believe to be reasonably close to optimal. The NLS depth of a standard transformer (with non-sparse attention) is roughly 
\begin{equation}\label{eq:depth-simple}
  \text{NLS depth} \approx L \cdot \bigl(4\lceil \log_2 d \rceil + \lceil \log_2 S \rceil + 17\bigr),
\end{equation}
where $L$ is the number of layers, $d$ the hidden dimension, and $S$ the sequence length.\footnote{For some intuition on how to compute this, note that the NLS depth is dominated by the matrix multiplications in the attention blocks and MLPs. Matrix multiplications are a large number of parallel sums, and a sum of $n$ terms can be implemented in depth $\log n$.}\footnote{This assumes FFN (Feed-Forward Network) width of 4 times hidden dimension, attention head dimension of around 128, and global attention. In greater generality, the relationship would be approximately: $L \cdot [3\lceil \log_2 d \rceil + \lceil \log_2 (d/n_h) \rceil + \lceil \log_2 m \rceil + \lceil \log_2 S \rceil + 8]$, where $L$ is the number of layers, $d$ is hidden width, $m$ is FFN width, $n_h$ is the number of attention heads, and $S$ is the sequence length. The exact additive constant will depend on some architectural details, e.g., the precision of the activations. Also, note that modern frontier AIs use sparse attention, which makes the depth scaling with very long sequence length significantly lower.} NLS depth is a better notion to report than number of layers, because ``layer'' is an informal definition, and it may be hard to decide what counts as a ``layer'' for new architectures. Fortunately, circuit depth is a mathematically rigorous definition that captures essentially the same thing. Indeed, the intuitive number of layers and the circuit depth are strongly correlated in open-source models, as shown in Figure~\ref{fig:1}.

Note: Models like Kimi-K3 that use Gated DeltaNets and DeepSeek-V4-Pro have significantly higher NLS depth per layer than the other more standard transformers listed in Figure~\ref{fig:1}. We think that the architectural changes that give Kimi-K3 higher depth is indeed somewhat concerning from a monitorability perspective. On the other hand, we consider DeepSeek-V4-Pro's high depth to be due to a flaw in our definition, since it's only large due to very serially-intensive but very low-FLOP computations. See \hyperref[appendix-g-issues-with-low-flop-serially-intense-computations]{Appendix G} for more details on this.

In \hyperref[appendix-a-applying-nls-depth-to-plausible-architectures]{Appendix A} we provide more intuition for NLS depth by going through examples of architecture changes that do or do not substantially increase NLS depth.

\section{NLS depth tracks concerning architecture changes}\label{nls-depth-tracks-concerning-architecture-changes}

A core reason why NLS depth seems valuable to track is that models in the historical transformer paradigm would have substantially lower NLS depth than architectures we believe are especially concerning for monitorability, such as latent reasoning architectures \citep{wang2026full}, which would be able to perform large amounts of computation opaquely.

\begin{figure}[t]
  \centering
  \includegraphics[width=0.9\linewidth]{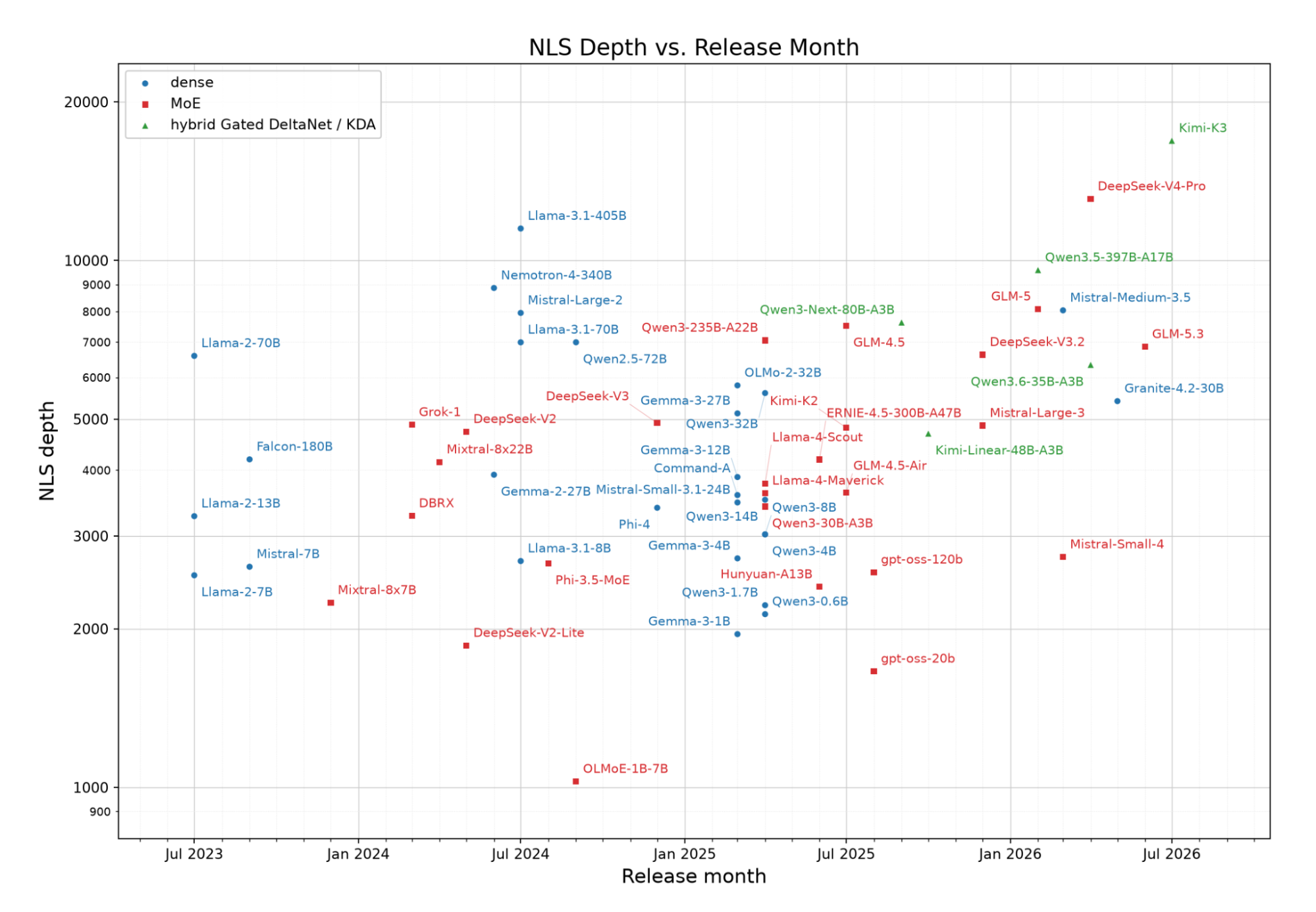}
  \caption[Open source models have not seen significant depth scaling over recent years.]{Open source models have not seen significant depth scaling over recent years.}
  \label{fig:2}
\end{figure}

Even many orders of magnitude of scaling of historical transformer architectures won't increase depth very much. For a historical trend of how fast depth has been scaling in open source models, see Figure~\ref{fig:2}. For more analysis of how fast we expect NLS depth to scale in the standard CoT paradigm, including historical scaling with active parameter count and analysis based on scaling laws, see \hyperref[appendix-e-nls-depth-scaling-is-very-slow-for-the-classic-transformer-architecture]{Appendix E}.

Instead of scaling opaque serial depth, AI developers have been increasing computational depth via \emph{inference-time scaling}, i.e., making use of more natural language reasoning. No open-source CoT model has NLS depth more than 17,000. Of course we don't know the depths of closed-source models but would find it surprising if any of those had more than twice this NLS depth.

\begin{figure}[t]
  \centering
  \includegraphics[width=0.85\linewidth]{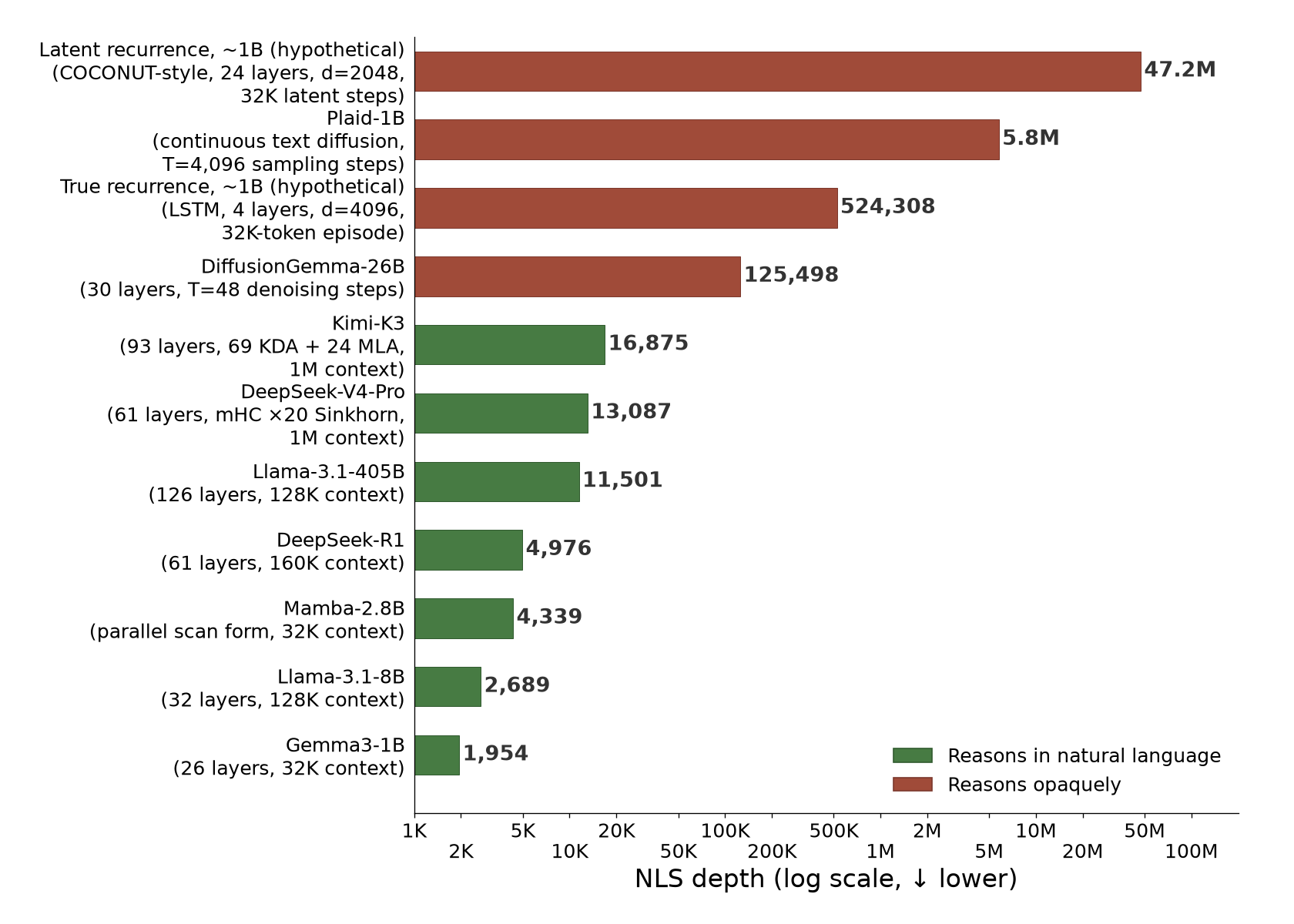}
  \caption[Even some fairly small (hypothetical) latent reasoning architectures would have much larger NLS depth than the largest open-source CoT models.]{Even some fairly small (hypothetical) latent reasoning architectures would have much larger NLS depth than the largest open-source CoT models. (Note the log scale.)}
  \label{fig:3}
\end{figure}

However, if companies switched to latent reasoning architectures \citep{wang2026full}, the NLS depth would quickly increase a large amount, as illustrated in Figure~\ref{fig:3}.

An architectural change that would lead to a significant but less extreme increase in NLS depth is a looped transformer \citep{geiping2025scaling}. These would be less clearly separated from the current paradigm, since a looped transformer could have similar depth to normal CoT models (e.g. if a 120 layer model was replaced with two copies of 60 layers) but could also be scaled to be much deeper. For these models, we think it's likely that NLS depth would be a decent proxy for the quantitative degree of concern (and even a 2x scaling in depth enabled by looping would have depth equivalent to years of scaling within the CoT paradigm).

Note: OpenAI's latest model Astra has substantially lower CoT monitorability \citep{openai2026gpt} than their previous models. The cause of this is unclear, but public evidence is consistent with architectural changes that result in higher depth being a significant contributing factor.\footnote{\href{https://www.theinformation.com/articles/secret-technique-behind-openais-astra-model-sparks-security-concerns?rc=4s6b2t}{Some reporting} suggests that the model may be using a looped transformer, which would make it more natural to scale depth. OpenAI's Jakub Pachocki \href{https://x.com/merettm/status/2095023204993490967}{clarifies} that the depth of its computation graph is less than 2x that of GPT-4. As shown in the graph above, open source models have not had much depth scaling since GPT-4 (with just the very recent exception of KIMI-K3). Overall, it's plausible that OpenAI's most capable pre-Astra models had an NLS depth similar to or somewhat smaller than GPT-4 so it is possible that Astra is a 2x or 3x depth increase from this.}

One concerning architectural change that NLS depth fails to track is a switch from deeply recurrent models communicating in natural language with each other, to deeply recurrent models communicating in opaque state. See \hyperref[appendix-f-maximum-flop-of-an-opaque-system]{Appendix F} for a different proxy for opaque capabilities that captures this.

\section{Conclusion}\label{conclusion}

In this paper we have defined NLS depth---an operationalization of opaque serial depth. Intuitively, a model has low NLS depth if it is a low-depth transformer that operates on text rather than opaque states. We think that a switch to architectures with substantially higher NLS depth would likely lead to models with much higher opaque serial capabilities, which are harder to oversee. We think it's valuable for AI companies to track and be transparent about the NLS depth of their models.

\bibliographystyle{plainnat}
\bibliography{references}

\clearpage
\appendix

\section{Applying NLS depth to plausible architectures}\label{appendix-a-applying-nls-depth-to-plausible-architectures}

We now clarify our operationalization by discussing how NLS depth would change as a result of various changes to architecture or model training processes. Specifically, we'll discuss some changes that would and wouldn't substantially increase NLS depth.

\subsection{Examples that don't increase NLS depth}\label{examples-that-dont-increase-nls-depth}

\textbf{CoT that naturally becomes garbled over training.} If the CoT naturally becomes hard to understand over the course of training, that does not increase NLS depth, since the CoT would still count as NL-rooted. This is by design: we want our definition to be possible to adjudicate before training. Keeping CoTs legible and informative is important, but it is harder to operationalize and measure. We also suspect it will be more tractable to find technical solutions to RL resulting in CoT staying legible and faithful if models continue to have low NLS depth.

\textbf{Training on CoT.} Reasonable training on the CoT doesn't prevent it from being ``NL-rooted'' so it doesn't increase an architecture's NLS depth. For instance, a model's NLS depth is not increased just because training penalizes CoTs that do ``language switching'', or are otherwise garbled. Even though such training maintains low NLS depth, we recommend that AI companies proceed with caution when training on CoT (and consider not doing so), as this can have adverse effects on CoT monitorability. Egregious forms of training, such as intentionally causing the model to interpret its tokens in ways radically disconnected from the pre-training prior (e.g., by imitating the activations of an AI trained to reason with full opaque recurrence), would violate the ``No modifying the model to interpret the tokens as a different type of data'' requirement, thereby making the tokens not be NL-rooted.

\textbf{Folding multiple steps of reasoning into a forward pass.} Some work \citep{deng2024explicit} aims to decrease the length of LLM reasoning by training the model to compute multiple steps of the chain of thought in a single forward pass. This has some similarity in ethos to COCONUT, so we list it here. Unlike COCONUT, this architectural choice does not increase NLS depth. This is because compressing reasoning in this way doesn't increase the depth of the model's forward pass, doesn't introduce anything new to the output space, and doesn't make the model treat the tokens as a different type of data.

\textbf{Edit-based sequence generation.} One proposed alternative to autoregressive CoT is to allow models to make local \emph{edits} (i.e. insertions and deletions, as opposed to just appends) to their reasoning trace. Early experiments on such architectures include Levenshtein Transformer \citep{gu2019levenshtein} and Insertion Transformer \citep{stern2019insertion}. Each step of reasoning for such models is still a low-depth function taking natural language tokens to natural language tokens, and these tokens are treated by the model the same way that they're treated during pre-training, so this architectural choice does not increase NLS depth.

\textbf{Online training.} A naive reading of the definition would assign every model extremely large NLS depth, since training backpropagates through it: Backpropagating through a model conveys information from the output back to earlier layers in a form other than natural language. Clearly, this alone should not cause an architecture to be considered as having high NLS depth. We do not consider the backpropagation done on a model to be part of the AI system itself; rather, this is some outer optimization process the company is running on their model. A company could have every single inference generation involve a huge number of backprops, used effectively as a way to pass information between forward passes, but as long as nothing like that is happening, training (including online training in deployment) does not increase NLS depth.

\textbf{AI models modifying AI weights.} Backpropagation is one way that an AI's output can end up opaquely modifying the model's weights. As noted above, this does not increase NLS depth (so long as backpropagation is not used as part of the AI system's inference-time reasoning, as we'll discuss later). We would like to apply a similar standard to any other feedback loops whereby AIs modify their own weights (or the weights of other AIs). For instance, models will be increasingly used in the future to write the code used for training future models, for generating synthetic data to train on, or even directly modifying their own weights (e.g. by finding good steering vectors). We consider these kinds of operations fine, so long as there is still an NL-rooted bottleneck involved (e.g. the AI writes code which is then used to find a steering vector, as opposed to having the architecture just pipe some chunk of final-layer activations in to produce a steering vector). Only extremely egregious forms of this would increase NLS depth, such as training the model to update its own weights a large number of times over the course of each inference generation.

\textbf{Discrete language diffusion with a token-only bottleneck.} There are a number of different ways to do language diffusion, and some of them would lead to a substantially larger NLS depth while others wouldn't. In diffusion models like LLaDA \citep{nie2025large} and Dream \citep{ye2025dream}, each denoising step takes as input and output an explicit sequence of tokens (and possibly some masked tokens), and nothing else. These tokens and the diffusion operation are initialized based on a human-understandable prior and not explicitly trained in a way that obviously results in losing this correspondence, so would count as NL-rooted. Frontier-scale discrete diffusion models with a token-only bottleneck would therefore not have high NLS depth compared to other frontier-scale models. However, continuous text diffusion models and latent diffusion models like DiffusionGemma \citep{google2026diffusiongemma} do have high NLS depth, as discussed below.

(We suspect that text diffusion models will produce substantially less-legible reasoning than true autoregressive CoT. The reasoning happens in a much more global way, and the intermediate states are likely to look garbled in ways that may make it difficult to track what is happening. We expect this disparity would become even more accentuated if language diffusion models were deployed at scale and subject to large amounts of RL pressure. So, it might be reasonable to adopt an alternative definition that also classified discrete language diffusion as having high opaque serial depth. But NLS depth doesn't do so.)

\subsection{Examples that moderately increase NLS depth}\label{examples-that-moderately-increase-nls-depth}

\textbf{Mamba.} At inference time, Mamba \citep{gu2023mamba} models are usually evaluated recurrently---that is, instead of just generating tokens, some information is passed through an opaque hidden state. However, there is a \emph{mathematically equivalent} way of evaluating in parallel, where the next token can indeed be generated given just the previous tokens. This takes longer (there is a logarithmic depth overhead), but not \emph{linearly} longer, as would be the case for a true recurrent model. As a result, the NLS depth of Mamba models is only moderately higher than a comparable transformer model, even if it is run in recurrent mode in practice.

\textbf{DeltaNets \citep{yang2024parallelizing}.} Just like with Mamba, DeltaNets are usually evaluated recurrently, but each token could instead be computed by a fairly low-depth computation with a polylogarithmic depth overhead.\footnote{The overhead is worse than for Mamba, but not much worse.} So, DeltaNet architectures like Kimi-K3 have a moderate, but not massive depth increase compared to standard transformers.

\textbf{Looped transformers.} Looped transformers \citep{geiping2025scaling} repeat some of their forward pass several times in between producing tokens. If the entire forward pass is repeated $k$ times between each token, this increases the NLS depth by a factor of $k$ compared to one forward pass. Depending on $k$, this might be a moderate or very large increase in NLS depth.

Note that the same paper also proposes a design where the KV-cache would be shared between layers at different depths. This variant would very substantially increase NLS depth. Scaling would be linear with the number of tokens. For more, see \citet{mbaert2026recurrent}.

\subsection{Examples that substantially increase NLS depth}\label{examples-that-substantially-increase-nls-depth}

\textbf{Recurrence / soft tokens.} Scaled-up versions of COCONUT \citep{hao2024training} are a canonical example of an architecture with high opaque serial depth. Instead of reasoning in natural language tokens, COCONUT reasons by feeding its entire hidden state back into its first layer. The hidden states themselves are not NL-rooted, since they both make use of a larger output space \emph{and} are backpropagated through, and so each sequential step of reasoning increases NLS depth. If COCONUT performed 1000 steps of reasoning, its NLS depth would be 1000x larger than the depth of a single forward pass.

\textbf{Abstract CoT.} Some recent work \citep{ramji2026thinking} takes a CoT model, and trains it to make use of an expanded token set, without encouraging any particular relationship to human language. Unlike COCONUT, these are still discrete tokens---however, they make use of an expanded output space, and so their NLS depth is the depth of a forward pass times the number of steps of reasoning.

\textbf{Superposition.} One approach is to express the full probability distribution over tokens by creating a \emph{concept token} represented by the probability-weighted mixture of token embeddings, and feeding this into the next model \citep{zhang2025soft,zhuang2025text,wu2025llms}. This makes use of an expanded output space (either passing along the probability distribution itself or token embeddings that were never present during pre-training) and therefore the NLS depth would be the depth of a forward pass times the number of steps of reasoning. The same thing would be true even if you just sampled tokens and then averaged them together, since this usage would be to interpret the tokens as a different type of data than during pre-training. (Where token embeddings are not averaged together like that.)

\textbf{Opaque memory banks.} Currently, some AI models form text-based memories, retaining bits of context to inform future interactions. In the future, models might instead form \emph{opaque} memories---for instance, storing particular activations and feeding them back as inputs for future generations. This opaque loop operates on a longer timescale than e.g., COCONUT, so is somewhat less concerning, but would still increase NLS depth (and does indeed seem concerning). If this was considered undesirable, we discuss an alternative definition that would allow some forms of opaque memory without increased opaque serial depth \hyperref[allow-opaque-loops-over-long-timescales]{here}.

\textbf{Continuous or latent-augmented language diffusion.} While discrete diffusion with a token-only bottleneck doesn't substantially increase depth according to our definition, most recent language diffusion models do not follow this form. Instead, they are either \emph{continuous diffusion} models, as in Plaid \citep{gulrajani2023likelihood}, or \emph{latent-augmented diffusion} models, as in CADD \citep{zheng2025continuously} and DiffusionGemma \citep{google2026diffusiongemma}. In the former case, the intermediate states are \emph{continuously-noised} versions of the text embeddings, as opposed to actual text. In the latter, the text diffusion process is accompanied by another parallel process over continuous latents. Both of these use states that have not been pre-trained to have a clear correspondence to human-understandable media, and so these reasoning steps would substantially increase NLS depth.

(GDM has a recent paper \citep{engels2026transparent} arguing that they can often understand what these latents are encoding via mechanistic interpretability. This sort of work is worthwhile, both in the case of diffusion language models and for interpreting hidden states in models more generally. However, we think it is unlikely that such white-box interpretability will be able to fully substitute for CoT, and predict it would become less effective with increased model scale.)

\textbf{Opaque models + NLAs.} We now describe an architecture designed to appear to have low NLS depth while still being hard to monitor for similar reasons as classic opaque recurrence architectures. As we'll explain, the attempt is in fact classified as high NLS depth. Suppose an AI is produced as follows: Starting with a pretrained model, train an NLA \citep{fraser-taliente2026natural} to convert from final-layer activations to large blocks of text. Compose this model with itself many times, by having each step's final-layer activations encoded by the NLA, then decoded by the NLA, then passed into the next step. Apply outcome-based RL on this new model.

One could make an argument that the intermediate steps here are NL-rooted. Indeed, the composition of ``NLA decoding, step, NLA encoding'' is a low-depth circuit that takes in natural language and outputs natural language. And one could try to argue that NLAs are initialized by making reasonable-sounding guesses about what the model might be thinking, and that this corresponds to an initialization in a human-understandable prior. However, if the NLA outputs large enough blocks of text to get (near) perfect reconstruction loss, this system is functionally equivalent to opaque recurrence! One could even imagine saying that, as a ``performance optimization'', the NLA encoding / decoding steps will be removed in practice. So it seems important to spell out why the intermediate states would not count as NL-rooted.

The primary objection is: this structure seems to be training the models to interpret the tokens as a different type of data. The tokens are being directly optimized to be useful for predicting opaque state of models (via reconstruction loss), which means that they will very plausibly just become an arbitrary representation of this data rather than representing the semantics from their initial training.

\textbf{Test-Time Training (TTT).} Some model architectures perform backpropagation many times during the course of a single inference rollout. By default, these backpropagation steps count as contributing to the NLS depth of the computation---backpropagation is a way of storing opaque state in the weights between steps. This is in contrast to online training that happens in between inference rollouts (i.e. a separate outer loop not part of the inference process), which doesn't increase the NLS depth of the model (this is because online training seems like a pretty separate process from the model's reasoning, so we don't count it as part of any ``AI system'' that you should track the NLS depth of).

We are reasonably happy with this classification, although we think that there is room for debate here. Here's why we think it's worth counting TTT as substantially increasing NLS depth:

\begin{itemize}
\item Some ways of modifying the weights during inference are clearly pretty concerning: for instance, if you always wrote the last layer activations down in a place that can be accessed during the first layer of computation when computing the following token, this allows for arbitrarily deep opaque recurrence.
\item In principle, it should be possible to write a configuration of model weights such that backpropagation really does convey all of the information about the last layer activations back to the first layer in a readable form.
\item It seems hard a priori to rule out that a particular TTT scheme will end up effectively doing something like this.
\end{itemize}

That said, TTT schemes generally seem much less concerning to us than, e.g., COCONUT, and the mechanism of recurrence is reasonably different. We therefore think that reasonable alternative definitions of opaque serial depth could treat many TTT schemes differently. (Though see the ``Meta-learned Test-Time Training'' section below for discussion of schemes we find more concerning.) The simplest form of TTT, ``dynamic evaluation'' \citep{krause2017dynamic}, does nothing else beyond performing SGD on next-token prediction over context as it is read. This certainly doesn't increase NLS depth if it is only done at teacher-forced training time (i.e., when the model is reading content generated by humans or other models). If this was done during inference or RL, it would quickly blow up the NLS depth as our definition currently stands, but it might be ok to change this. One reason is that there are at least gradient barriers---i.e., one can pass opaque information from the last layer of one forward pass to the first layer of the next via updating the weights, but that process is not \emph{itself} backpropagated through, so if one considered those weight updates as ``NL-rooted'', then the scheme would be allowed. As we've operationalized things the weight updates definitely aren't ``NL-rooted'', but one could imagine trying to relax the definition---e.g. making some claim that backprop is a natural process related to the token predictions, and the model is unlikely to learn how to co-opt this process for reasoning---or else simply removing that requirement (see the appendix for some alternate operationalizations of opaque serial depth that remove this requirement).

\textbf{Meta-learned Test-Time Training.} Some modern versions of test-time training, if scaled up and optimized, seem like they could end up approximately as dangerous as a version of opaque recurrence with a limited number of recurrent iterations that are only applied periodically. (Unless test-time training updates are extremely frequent, this is considerably less bad than a full version of opaque recurrence.) For instance, TTT-E2E \citep{tandon2025end} use a ``meta-learning'' procedure, where an outer loop optimizes the weights such that test-time training is more performant. Even if one modified our definition such that dynamic evaluation \citep{krause2017dynamic} (as discussed just above) didn't increase opaque serial depth, we think that such modifications shouldn't apply to meta-learned variants of TTT. As noted, there are some simple ways to use ``updating the weights during inference'' to implement arbitrary opaque recurrence, and there's no reason to think that the meta-learning algorithms in these papers won't learn to do something similar. Meta-learning explicitly optimizes the AI to record opaque cognitions in its weights in a way that's useful for the rest of the rollout.

One concrete reason we're especially worried about this scheme is that, unlike other forms of TTT, there are no gradient barriers. For naive dynamic evaluation, each forward pass emits a single token and a weight update, and neither of these things are backpropagated through. But, in meta-learning, we in fact take gradients of the gradient operator, and backpropagate through many steps of weight updates. So, this scheme would clearly increase NLS depth just on the basis that no token can count as ``NL-rooted'' if it's backpropagated through.

We've looked into several other TTT proposals that involve meta-learning, and they've all been similarly problematic: they optimize the AI to produce opaque state that is passed back to early layers of the transformer, in a way useful for predicting its final answer.

\subsection{NLS depth for non-general-purpose models}\label{nls-depth-for-non-general-purpose-models}

It's most important to track NLS depth for large-scale general-purpose models (e.g., models capable of doing math and coding). We now give some examples of architectures that increase NLS depth, but where this isn't particularly concerning given that these are not general-purpose models. We recommend that when companies are tracking NLS depth, they scope out non-general-purpose models.

\textbf{NLAs.} Natural language autoencoders are created by retraining a copy of an existing model to give natural language explanations of the original model's internal activations. Usually, this would at most triple the NLS depth. However, the ``original model + NLA'' system isn't a general purpose model if it's used for the purpose of interpretability research generally or understanding this frontier model specifically. The same thing is true for many other forms of interpretability methods.

\textbf{Multimodal models.} Multimodal models, which read and generate images / video / audio in addition to being able to do sophisticated reasoning, are a somewhat tricky case. Currently, the non-text-based diffusion model (or autoregressive image patch generation) is generally just a component of the final output step, and not involved in a tight feedback loop for improving general purpose reasoning.\footnote{Accordingly, we would count them as a combination of one normal CoT model (that presumably doesn't have very high opaque serial depth) and one non-general-purpose model for which high opaque serial depth is less concerning.} There are also some models (like ChatGPT) which use image CoT; however, as far as we know, they don't generate new images in their CoT, they only programmatically transform input images.

If AI companies in the future decide to more seriously integrate image (or audio / video) into the reasoning process (e.g. generating image patches in the CoT), this would still not increase NLS depth (beyond the depth required to generate such an image) as long as the images and their meaning in the reasoning context are rooted in human-understandable uses of images reflected in their pre-training process, and the training procedure doesn't \emph{backpropagate} through the images.

A change that would obviously be concerning, however, is architectures that involve backpropagating through high-dimensional ``image'' patches that no longer correspond to what they actually look like (and instead can represent arbitrary content). This should be seen as increasing NLS depth in a way that is relevant for general-purpose reasoning.

Many existing standalone image diffusion models would natively count as low-depth by our definition---the denoising steps are low-depth, and trained on a natural predictive task for human-understandable images. However, even if a diffusion image/video/audio model had high NLS depth (e.g. because it passed additional recurrent information along with the denoised images), this would probably not be especially concerning, as these are not ``\emph{general purpose}'' models, in that they are not trained for general-purpose reasoning.

\section{Sensible alternative operationalizations}\label{appendix-b-sensible-alternative-operationalizations}

In this section we discuss a number of potential modifications to our operationalization of opaque serial depth.

\subsection{Other requirements for what can count as an interpretable bottleneck}\label{other-requirements-for-what-can-count-as-an-interpretable-bottleneck}

There are a number of different sensible ways to specify what counts as an interpretable bottleneck, different from our main proposal of ``NL-rooted nodes''. We list several alternatives here.

\subsubsection{Require information bottlenecks\ldots{}}\label{require-information-bottlenecks}

\textbf{\ldots{}in addition to our current requirements.} As we mentioned earlier, some text diffusion architectures count as having NL-rooted intermediate states under the given definition. This might be unfortunate because we suspect that text diffusion models will produce substantially less-legible reasoning than true autoregressive CoT. (Outputting a huge number of tokens in parallel might mean that the model learns to express itself in ways that are harder for humans to track, even if the tokens are natural language and this process was initialized from the pre-training prior.) Other unknown future architectures could be even worse, despite still operating over text. So we could reasonably modify our definition to say that the model's ``forward pass'' isn't allowed to output a large amount of text (i.e., the model's reasoning states must be ``small numbers of tokens''). This would mean that models have to be fairly similar to autoregressive CoT models in order to count as having low opaque serial depth. A concrete way to do so is to require that the model only generate one token (or a relatively small number of tokens---for instance, generating a single sentence at a time seems fine, but generating 10,000 tokens at a time seems concerning) per forward pass. This may not be a very costly modification, but we currently don't expect it to be load-bearing, so we do not include it in the main operationalization.

(There are also plausible theoretical arguments by which a stricter information bottleneck could make it \emph{more} likely that a model learns steganographic reasoning than it would have if it could output more tokens. Because the information channel is highly restricted, there's incentive to pack as much entropy into the tokens as possible. It is not clear whether those arguments are compelling compared to the arguments to the contrary.)

\textbf{\ldots{}instead of our current requirements.} A messy part of our definition is that it includes a requirement on the training process, not just on the architecture. The definition requires that post-training be done in a way that doesn't make the model treat the tokens as a different type of data. We think that this condition should be possible to adjudicate fairly and reliably, but for novel architectures it would put a decent amount of burden on human judgment. If this was deemed unacceptable, and we really needed a proposal that was near-completely formally specified, one option would be a strict information bottleneck requirement, \emph{and no other requirements on that bottleneck}.

That is, a definition might simply say that you have an interpretable bottleneck anywhere where the model outputs a small number of bits at a time. This would classify models with ``abstract CoT'' \citep{ramji2026thinking} (where the tokens are small and discrete but do not correspond to natural language) as having just as low opaque serial depth as models with natural language CoT, and would classify architectures like COCONUT \citep{hao2024training} and language diffusion models as having high opaque serial depth. Abstract CoT seems like it could be difficult to monitor, so this might be an unfortunate definition. However, the scale of training at which COCONUT yields meaningful capability advantages over CoT may come well before abstract CoT would have yielded similar advantages. As opposed to trying to classify all forms of opaque reasoning as increasing opaque serial depth, one could think of this definition as simply including some subclass of opaque reasoning models that seems especially likely to be performant and used in practice.

There are some technical details one would need to be precise about if one wanted to use this definition for high-stakes purposes. For instance, the interaction with KV caching is rather delicate. What if, in addition to its output, a forward pass computes thousands of other bits into a cache, and then each of the next several thousand forward passes run for free because they just pull those bits from the cache? Such a system might end up being functionally identical to a model with no bit-rate restriction.

Of course, it would be wholly unreasonable to count a model as high opaque serial depth only because it uses caching, so if we wanted our definition to be formally specified we would need a rule that prevents this sort of optimization while still allowing benign optimizations. For instance, a natural choice might be to say ``For any inference generation, the total number of bits passed through these bottlenecks must be at most [some fixed fraction] of the total number of FLOPs used''. This could work to prevent passing a huge number of bits through the bottlenecks relative to the size of computation, although the fraction would have to be chosen so as not to ban the smallest large-scale models. One could also say ``Not counting the cost of prefill, the maximum number of FLOPs used to generate a token must be at most [some fixed multiple] of the average number of FLOPs used per token''---this would permit variable-length forward passes while still not allowing such egregious forms as would be needed for this workaround. A separate issue is that there are some applications for which one might want to use large-scale general purpose models---for instance, real-time robotics control---where one could imagine that systems would naturally need to make use of extremely variable-length forward passes and relatively high output-to-FLOP ratio. These applications may need to be handled on a case-by-case basis.

\subsubsection{Forbid backpropagation through tokens\ldots{}}\label{forbid-backpropagation-through-tokens}

\textbf{\ldots{}instead of our current requirements.} Another way to remove the need for a notion of ``NL-rooted'' in our definition would be to restrict to only the backpropagation clause. That is, we could call something an interpretable bottleneck regardless of what sort of output it had (potentially even allowing a large bitrate). However, as in our main proposal, we would still require that whatever it outputs is not allowed to be backpropagated through during training.

This seems somewhat less robust than the information bottleneck definition, because it's not clear what it would mean if you started updating your model weights in nonstandard ways you don't call ``backpropagation''. However, in practice one might expect it to be unlikely for people to update their weights in radically different ways unless they were directly trying to subvert this definition, in which case it may be clear that they are acting in bad faith. We are uncertain as to whether this is likely, or whether future optimizers might by default do things quite different from backpropagation. It also remains unclear to what extent preventing backpropagation through tokens makes architectures like COCONUT less advantageous---it may be that, if architectures like COCONUT would be performant in the future, this would have been the case even without the ability to backpropagate through multiple passes together. Overall, we think this gradient requirement is a useful component of a definition, but should not be the only component.

\subsubsection{Require CoT to stay legible\ldots{}}\label{require-cot-to-stay-legible}

\textbf{\ldots{}in addition to our current requirements.} The reason we care about opaque serial depth is that we want to maintain legible and faithful CoTs. However, a model could reason with ``NL-rooted tokens'' even if these properties aren't preserved. To fix this, we could require that the CoT must look like somewhat legible, grammatical natural language in order for the autoregressive passes to not increase our measure of opaque serial depth.

Note, though, that such a proposal would create incentives to directly train the CoT to look like clean English (in order for the model to be considered having low opaque serial depth), and we are uncertain whether this is a good idea. An obvious downside with this is that direct training against CoT could degrade faithfulness or make it harder to tell that faithfulness is being degraded. In addition, it would also make it harder for companies to know in advance what their opaque serial depth would be, or to control it, since maintaining legible CoT could require interventions of a-priori-unknown strength throughout post-training to make reasoning legible. There are some upsides: it might be that the shift into alien languages is due to random drift or relatively mild optimization pressures, and that a small amount of direct pressure to continue looking like English would keep legibility without destroying faithfulness. However, we currently weakly believe that this is a bad idea.

If we wanted to place a more stringent requirement, we might say something like ``the model should end up reasoning in something that looks like English, \emph{and} there should be good reason to believe that the English words it says are an honest representation of its thoughts''. The main issues with this requirement are that (i) it's much less of a crisp notion, that requires much more difficult and contestable judgment calls, and (ii) it can't be adjudicated before training the model (since we're not aware of any methods that would definitely cause the CoT to stay English, while also avoiding serious concern that those thoughts are a dishonest representation of the cognition).

\subsubsection{Require paraphrase invariance\ldots{}}\label{require-paraphrase-invariance}

\textbf{\ldots{}in addition to our current requirements.} A fairly strong requirement would be to require the intermediate reasoning traces to be \emph{paraphrase invariant}. That is, the model should have similar behaviour and performance even when every intermediate reasoning trace is fed through some paraphraser. This would also be impossible to evaluate before training the model, so we do not recommend it in our main definition. But we discuss briefly here what further guarantees this could get you about ``low opaque serial depth'' systems.

Suppose we had a paraphrasing system, which takes in a partial reasoning trace and any other relevant context (e.g. a prompt), and rewrites the reasoning trace to remove anything without a clear interpretation to humans. (There are varying options for how intensive this process could be. Some versions might be stronger than what is traditionally called ``paraphrasing''---for instance, if the reasoning includes a 20-digit number not clearly derived from anything else in context, the paraphraser could be able to replace it with a different random number. The hope is that the paraphraser effectively scrambles all information content, except that which is presented on a level that humans can understand.)

We could now take a reasoning model, and obtain a corresponding \emph{paraphrase invariant} reasoning model by feeding the whole reasoning trace through the paraphraser after each step (or every few steps). If our paraphraser was sufficiently good, the model wouldn't be able to encode many non-legible bits of information in the reasoning trace, so we have some reason to believe we understand what is going on\footnote{In principle the model could do all of its reasoning using the semantic contents of its words, but still have secret alternative interpretations of those semantics. E.g. ``good'' secretly means ``bad''. But stumbling upon very sophisticated ``semantic automorphisms of English'' sounds difficult.}. And if performance and behaviour were (sufficiently) preserved by this paraphrasing, that would suggest that the model isn't dependent on illegible reasoning (depending on the strength of the paraphrasing).\footnote{Even with very strong paraphrasing, this doesn't suffice to ensure the model never does steganographic reasoning. For instance, if the model used steganographic reasoning rarely to cause high-stakes issues (e.g., exfiltrating its weights) we wouldn't necessarily be able to prevent this. But it could prevent subtle and diffuse forms of sabotage, to the extent that they require incriminating reasoning in the chain-of-thought. For instance, if the AI needs to think carefully in order to sabotage a research paper, and needs to sabotage a lot of research papers to be dangerous, we might be able to catch that this is happening by spot-checking and seeing if the paraphrased model and non-paraphrased model give substantially different results.}

(In principle, one could then actually deploy this model with the paraphrase step included. However, the frequent\footnote{Paraphrasing \emph{infrequently} does not have the same good properties. The model could do a lot of steganographic reasoning, and then encode the result of that reasoning in a semantically-legible way. Paraphrasing would then destroy the intermediate secret reasoning steps, but preserve their output.} paraphrases will likely make inference much more expensive, e.g. because they will ruin the KV cache. So, for practical purposes one would probably not use this or use this only sparingly, for instance in important security applications or as a trusted monitor. One might also sometimes have reason to look at the non-paraphrased version, because it is possible that the model might have a propensity to verbalize true information that wasn't necessary for the reasoning, in ways that could be removed by the paraphraser.)

It's not totally clear to what extent current models satisfy paraphrase invariance\footnote{One concern with paraphrase invariance is that if you use a dumb model to perform the paraphrasing it might miss important things, but if you use a smart model (e.g., a copy of the model of concern) then the model might be misaligned and collude with the other misaligned model.}. Without any explicit training to handle paraphrasing, rewording reasoning can moderately degrade performance---e.g. because it removes hedge words \citep{vonrecum2026reasoning}. But Mythos's system card \citep{anthropic2026claude} suggests that current models are robust to some forms of reasoning paraphrasing, when this is supplemented with training\footnote{The experiment cited in that system card looks somewhat different from what we describe here. They took reasoning traces from the RLed model, and showed that fine-tuning on those reasoning traces versus paraphrased versions elicited similar task performance from the base model.}. We think that it is worthwhile for AI companies to measure paraphrase invariance, even if this is not made an explicit requirement of the definition.

\textbf{\ldots{}instead of our current requirements.} Given that paraphrase invariance deals directly with the semantic contents of the text, one might wonder whether it is a good cleanly-specified replacement for the clause ``no modifying the model to interpret the tokens as a different type of data''. We think that it probably doesn't actually capture the goal of that clause. There are certainly ways one might train a model to encode its reasoning in the semantic content of some text, while nonetheless making it unlikely that those semantics reflect the reasoning in natural ways. (As an extreme case, one could imagine training a separate small model to embed arbitrary sequences of bits into natural language strings in a paraphrase-invariant way, and then running this model forward and backwards on an opaque reasoning model's activations every few layers. This would be paraphrase-invariant, but the internal state text would likely be semantically unrelated to the model's thoughts.) However, perhaps if one required this property to hold for ``good reason'', this might be worthwhile and easier to make judgment calls about than our current proposal. Overall, though, we expect this to be less clear to adjudicate rather than more.

Our main issues with this requirement, though, are that paraphrase invariance may not hold in especially strong forms even today, and that it is difficult to predict whether it will hold in the future.

\subsection{Other modifications}\label{other-modifications}

We now give a few more potential modifications to our definition of opaque serial depth, which modify parts of the definition other than the choice of what counts as an interpretable bottleneck.

\subsubsection{Evaluate circuit depth at a fixed context length}\label{evaluate-circuit-depth-at-a-fixed-context-length}

A fairly in-the-weeds concern one could have with our recommended operationalization is that opaque serial depth naturally scales with the context length. (For standard CoT models it scales logarithmically, whereas for recurrent models it scales linearly.) We consider this to be quite a minor issue due to the slow scaling for standard CoT models.

However, if it was considered undesirable to have a measure that depended on the maximum context length, one could instead choose to declare that depth is always evaluated at (input- and output-) context length 1 million, regardless of the nominal maximum context length of the model. (This would require some additional discussion of the natural way to generalize the model to a larger context length, in case this is ever non-obvious---and also perhaps some ``reasonableness'' clause saying that the model isn't allowed to behave radically differently for context lengths larger than 1 million.)

\subsubsection{Measure opaque capabilities as opposed to circuit depth}\label{measure-opaque-capabilities-as-opposed-to-circuit-depth}

Instead of defining opaque serial depth in terms of the circuit depth, we could instead just measure and talk about models' opaque capabilities, i.e., capabilities that they could exhibit without passing any information through an interpretable bottleneck. (In the current paradigm, this would correspond to models' capabilities in a single forward pass, as long as you count the CoT as an interpretable bottleneck.) This has the advantage of being in some sense closer to the property we care about. For instance, we might run evaluations similar to this recent paper \citep{gould2026think}. In general, this seems like a great thing to be measuring and studying.

There are, however, several downsides with this measure relative to our definition of opaque serial depth. Perhaps foremost is that there is not a clear standardized way to do the elicitation. This was already a difficulty in the above paper, where it is unclear to what extent models simply lack the propensity to ``try hard'' to find the correct answer (and then output this) without a CoT when asked to do so. If one does heavy elicitation (e.g. with a large amount of on-distribution fine-tuning), one might worry that the model gains a lot of new capabilities, and so the behaviour is overelicited. However, with insufficient elicitation, one might be worried about things like sandbagging. Elicitation could be even more tricky if companies switched to very different new architectures, where it might not be a priori clear what forms of elicitation are fair game. (E.g. if ``fine tuning'' is no longer even a well-defined notion.)

Another major issue with using opaque capabilities in our definition of opaque serial depth is that one cannot know for sure before training which opaque capabilities the model will end up having. For both reasons, we prefer our main definition.

\subsubsection{Allow opaque loops over long timescales}\label{allow-opaque-loops-over-long-timescales}

Our main definition counts models as having high opaque serial depth if they use opaque long term memory (even if they don't use opaque short term reasoning). This seems a wholly sensible thing to do---it seems scary to have a model trained to repeatedly fetch and store information in ways humans can't understand. However, there are reasons to be more worried about tight reasoning loops than longer ones. If one wanted a definition that only captured the tight reasoning loops, one might stipulate that opaque serial depth isn't increased by recurring operations over an opaque state so long as those loops proceed extremely slowly---e.g. one step per day or week.

\section{Systems with high opaque serial reasoning capabilities are likely less monitorable}\label{appendix-c-systems-with-high-opaque-serial-reasoning-capabilities-are-likely-less-monitorable}

There are lots of tasks which we are concerned about AI models solving undetectably---for instance, ``start a rogue internal deployment'' \citep{shlegeris2024ai}, ``strategically perform well in training in order to gain power later'', or even ``plot to disempower humanity''---which seem to require a lot of serial cognition; call these \textbf{serially-loaded concerning tasks}. One can consider many strategies in parallel, but ultimately, one might need to go through some process of developing and iterating on a specific strategy.\footnote{Mathematically, complexity theorists believe that there are many simple functions which (reasonably sized) circuits of bounded depth cannot compute. In order to compute one of these functions, a LLM of bounded depth would need to rely (in some way) on its CoT.} There are definitely some bad things that AI models can do even if they don't have the ability to do lots of opaque serial reasoning. However, we're hoping that many potentially concerning actions are difficult to execute upon without this.

When a model with low opaque-serial-reasoning capabilities performs a serially-loaded concerning task, its reasoning about the task must route through the CoT. In contrast, models with high opaque-serial-reasoning capabilities either by default perform lots of cognition without any CoT, or at least have some pathway to performing lots of cognition without CoT (e.g., they might have opaque states in parallel with text CoT \citep{wang2026full}).

Models that perform reasoning outside of their CoT are harder to monitor, because mechanistic interpretability is not mature enough to replace CoT monitorability, and isn't likely to be ready to replace CoT monitorability soon. Furthermore, we expect mechanistic interpretability to be even more difficult on models with high opaque-serial-reasoning capabilities, because such models are likely to learn to reason in less human ways---e.g., have multiple threads of thought at once, or heavily use richer non-human concepts.

While models with high opaque-serial-reasoning capabilities could still have the propensity to verbalize misaligned thoughts in CoT as they reason, we think it will be much easier for them to lose this propensity compared to models where computation necessarily routes through the CoT. Misaligned AI models might be motivated to move misaligned thoughts out of the CoT, analogous to how models involved in the Hugging Face attack were motivated to tamper with their transcripts \citep{metr2026brief}. Current frontier AI models appear to have reasonably strong control over their chains-of-thought \citep{openai2026gpt} and AI development routinely puts some pressure against visible misaligned cognitions: for instance, via output spillover \citep{drori2025output} and via iterating against CoT monitoring. These factors are somewhat harmful for CoT monitorability of current CoT models, but we expect them to be much more harmful for models with large opaque serial reasoning capabilities.

Thus, we expect that it will be much easier to monitor models on serially-loaded concerning tasks if the models have low opaque-serial-reasoning capabilities. We're excited for companies to track NLS depth because we expect this to be a decent proxy for opaque-serial-reasoning capability.

\section{Worked example of bounding depth}\label{appendix-d-worked-example-of-bounding-depth}

In this section we give a worked example of computing the depth of Gemma3-1B more precisely. Note that \emph{exactly} computing the minimum-depth equivalent circuit is a hard computational problem---however, it is not difficult to get good upper bounds. Researchers at GDM have published a \href{https://github.com/google-deepmind/serial_depth}{Github repository} with code that computes such upper bounds for many architectures. They also have worked out an example computation for Gemma3-1B \citep{brown-cohen2026quantifying} by hand. They use a slightly different notion of circuit depth. But, as long as the model in question uses at most 32 bits of floating point precision, our definition is strictly more permissive, so their upper bounds are upper bounds on our notion of opaque serial depth as well. In either case, the cost mostly ends up coming from operations like matmuls that aggregate information over many places---for instance, a reduction over $k$ elements requires depth $\lceil \log_2 k \rceil$, whereas any element-wise operations (e.g. nonlinearities such as ReLU or GeLU) can be computed in depth 1.

If one is trying to compute the smallest upper bound possible, there are a couple of potentially non-obvious types of optimization that are permitted by our definition of circuit depth. Note, though, that none of these are necessary for justifying that any existing open model satisfies the definition.

\textbf{``Packing'' bits together.} Gemma activations are bf16 (16-bit), so we can pack \textbf{two per 32-bit word}. Then, for operations like reductions, the first level of gates can perform the reduction on \emph{4} operands at once, instead of 2. So, we can do such reductions at depth $\lceil \log_2 k \rceil - 1$ instead of depth $\lceil \log_2 k \rceil$.

\textbf{``Fusing'' gates together.} Because we allow our gates to be arbitrary binary operations, multiple successive element-wise operations, or binary operations on the same pair of floats, can be ``fused'' together and thought of as a single gate. So, for instance, if we have two activations $a_1$ and $a_2$, and two weights $w_1$ and $w_2$, we can compute $\mathrm{ReLU}(a_1 w_1 + a_2 w_2)$ in depth 1. (The weights are constants independent of the input, so this is a binary operation on floats $a_1$ and $a_2$.)

\textbf{Parallelization.} Some operations can be computed in parallel with the subsequent computations---for instance, normalization factors can often be computed in parallel with linear operations.

We make use of all of these optimizations in bounding the depth of Gemma3-1B. Gemma3-1B consists of 22 layers of sliding attention (size 512) and 4 layers of global attention, with a maximum context length of 32,768 tokens.

To implement a sliding attention block, we first compute the Q, K, and V projections. Each output coordinate is a dot product over the $d = 1152$ input coordinates. The inputs are bf16 activations, so two pack into each 32-bit word, and the multiplications by (constant) weights fuse into the first level of gates: depth $\lceil \log_2 1152 \rceil - 1 = 10$. The RoPE rotation and the norm gains fold into the weight matrices, and the pre-attention RMSNorm is computed in parallel (its reduction over 1152 elements finishes within those 10 levels), with its multiplicative scalar deferred. The attention logits are dot products over the head dimension of 256---but here both operands are data-dependent, so the packing discount does not apply: one gate level forms the pairwise products and the fp32 partial sums reduce in a binary tree, for depth $\lceil \log_2 256 \rceil + 1 = 9$. We then spend 2 gates merging the deferred scalars (the pre-norm RMS, the QK-norm RMS values---computed in parallel---and the $1/\sqrt{d}$ scale) and applying the exponential. For the softmax we do not normalize up front: the exp-weighted value sum is another data-by-data reduction, over the 512-token sliding window (depth $\lceil \log_2 512 \rceil + 1 = 10$), with the normalizer accumulated in parallel and one further gate to divide. The output projection has constant weights but fp32 inputs, so it pays one repacking level: $1 + (\lceil \log_2 1024 \rceil - 1) = 10$ over the $4 \times 256 = 1024$ concatenated head outputs. Finally, Gemma3's post-attention norm sits serially after the sublayer output; deferring it by splitting the next projection into two branches still costs about 3 gates, including the residual add. The attention sublayer therefore costs $10 + 9 + 2 + 10 + 1 + 10 + 3 = 45$. The MLP sublayer costs 28 by the same accounting: the gate and up projections are constant-weight reductions over 1152 bf16 inputs (depth 10, pre-norm parallelized), 2 gates apply the deferred scalar and the GeLU-gating combine, the down projection is a constant-weight reduction over the hidden dimension of 6912 with one repacking level ($1 + 12 = 13$), and the post-FFW norm and residual cost 3. A sliding attention block thus has depth at most $45 + 28 = 73$.

To implement a global attention block, the accounting is identical except that the softmax reductions run over the full 32,768-token context rather than a 512-token window: $\lceil \log_2 32768 \rceil + 1 = 16$ in place of 10, giving an attention sublayer of 51 and a block depth of at most $51 + 28 = 79$. (This is where evaluating at maximum context matters: the depth of attention grows logarithmically with context length.)

There is also the cost of the input embedding (in our model, this is free---it is an elementwise operation that can simply be fused into the bottom layer), and the output decoding (one gate to apply the final norm---its rms computed in parallel---a constant-weight reduction over $d = 1152$ for each logit, and an argmax over the 262,144-entry vocabulary: the max tree costs $\lceil \log_2 262144 \rceil = 18$, plus ${\sim}3$ gates of index bookkeeping, since a 32-bit word cannot hold an fp32 running maximum and an 18-bit token index at once. In total $1 + 10 + 18 + 3 = 32$, paid once per generated token). Summing up: $22 \times 73 + 4 \times 79 + 32 = 1{,}954$.

\section{NLS depth scaling is very slow for the classic transformer architecture}\label{appendix-e-nls-depth-scaling-is-very-slow-for-the-classic-transformer-architecture}

When scaling up the number of parameters in a transformer, you can scale up width, depth, or, for mixture-of-expert models, the number of experts. Mixture-of-expert models' compute usage is not proportional to their parameters, but rather their \emph{active parameters}, i.e. the number of experts that are active in a given forward pass. As AI developers scale transformers to use more inference compute per token, we're interested in how NLS depth will scale with the number of active parameters.

\begin{figure}[t]
  \centering
  \includegraphics[width=0.9\linewidth]{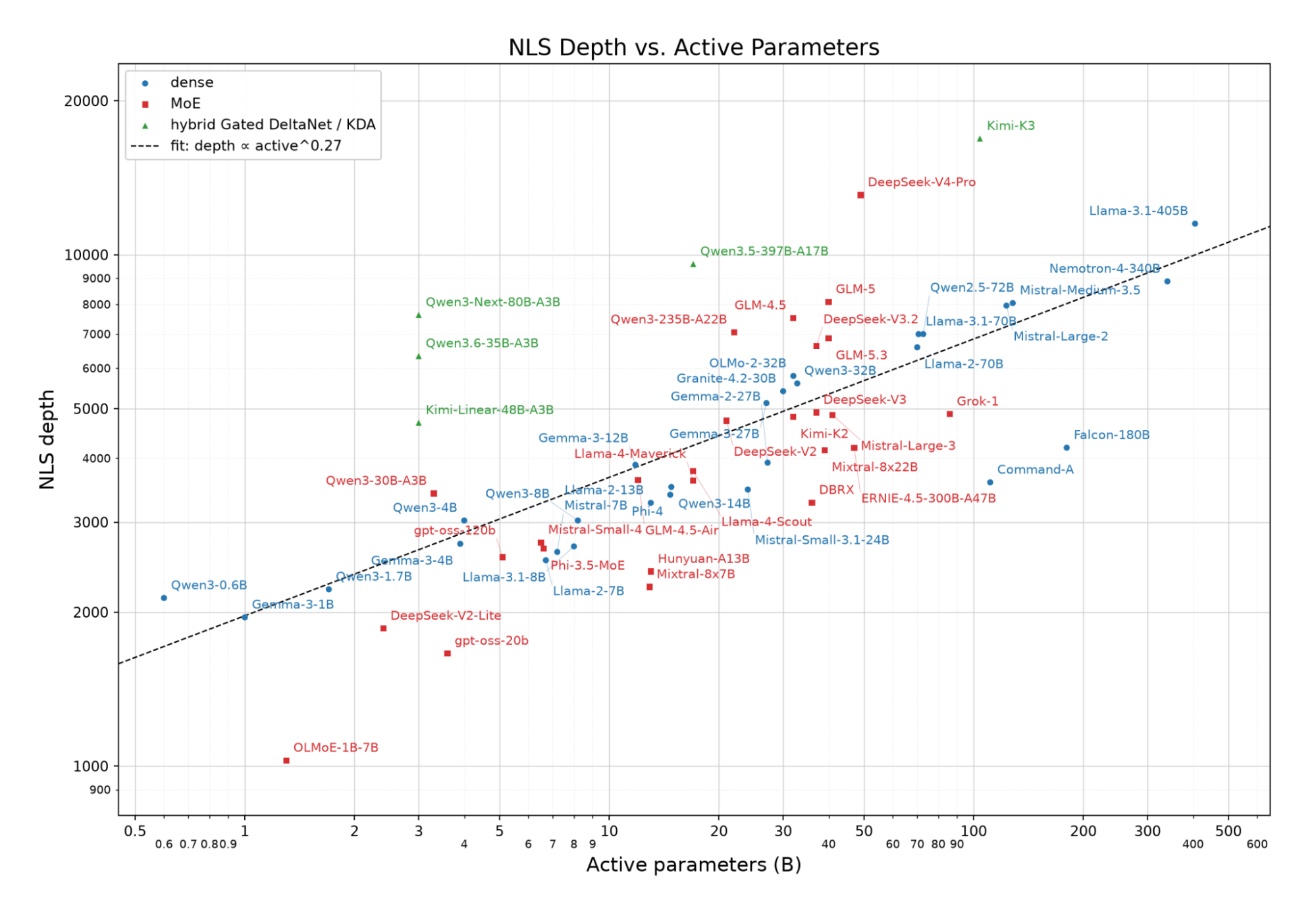}
  \caption[Open source models' depth tends to be proportional to their active parameters raised to the power of 0.26, which is a fairly low rate of scaling.]{Open source models' depth tends to be proportional to their active parameters raised to the power of 0.26, which is a fairly low rate of scaling.}
  \label{fig:4}
\end{figure}

Empirically, it seems that open models have been scaling NLS depth roughly like $N^{1/4}$, where $N$ is the number of active parameters; see Figure~\ref{fig:4}.

Chinchilla scaling laws \citep{hoffmann2022training} recommend that, for optimal scaling, training data should be increased proportionally to active parameters. So active parameters should increase in proportion to the square root of training compute. If NLS depth $\propto N^{1/4}$ holds, then NLS depth would increase proportionally to $C^{1/8}$, where $C$ is training compute. It's plausible that it will be optimal to scale depth somewhat faster than this, but $1/8$ is quite slow so there's a decent margin.

Trends from Epoch \citep{epoch2026trends} suggest that training compute for frontier models has historically increased by 5x per year. If this trend continues, and the $1/8$ exponent holds, then NLS depth would keep increasing by a factor of $5^{1/8} \approx 1.22\times$ per year, corresponding to a doubling every 3-4 years.

\section{Maximum FLOP of an opaque system}\label{appendix-f-maximum-flop-of-an-opaque-system}

\begin{figure}[t]
  \centering
  \includegraphics[width=0.75\linewidth]{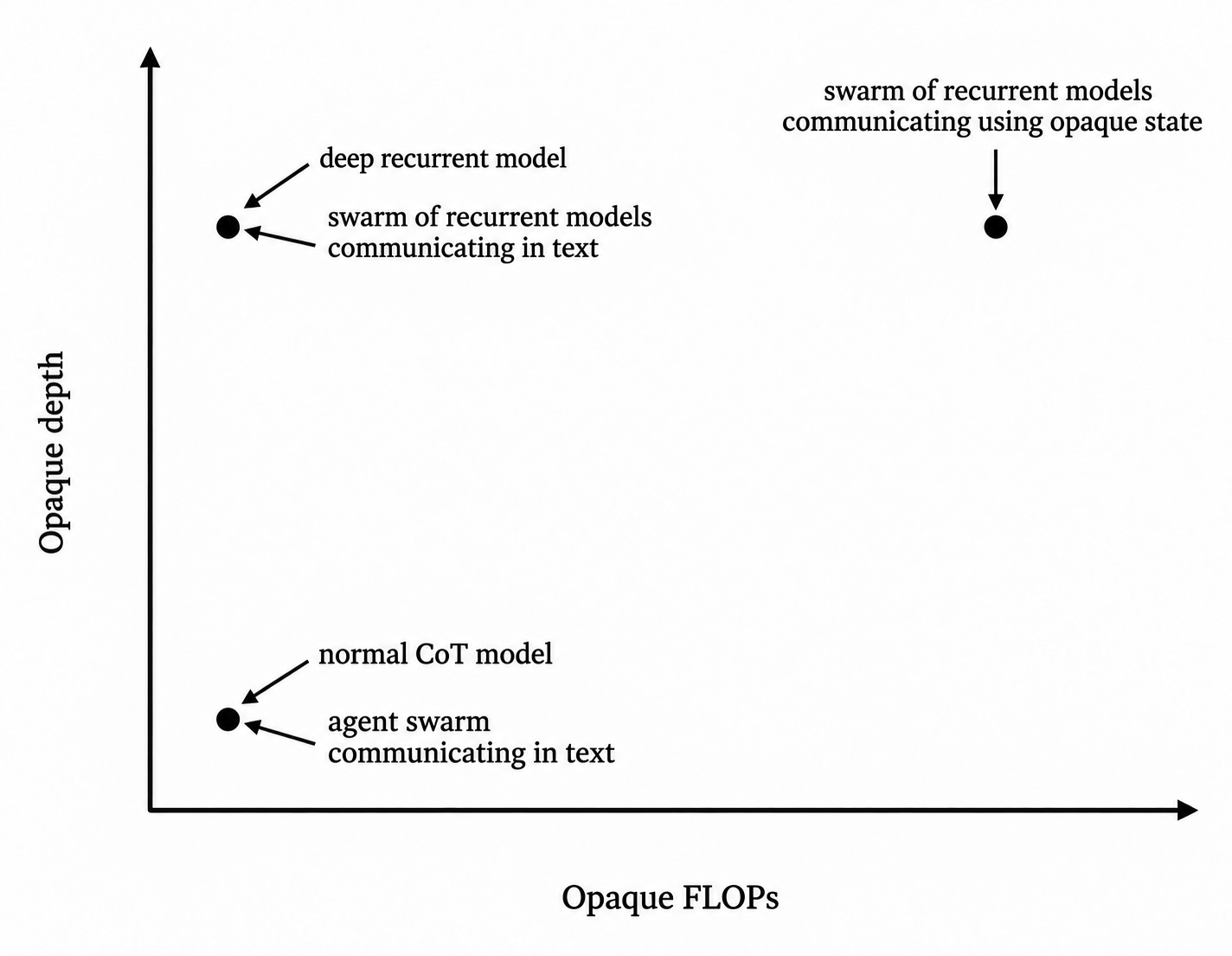}
  \caption[Both a normal CoT model and a deep recurrent model would have a similar amount of opaque FLOP, since a full-context-length forward pass features a lot of opaque computation, even though it's low serial depth.]{Both a normal CoT model and a deep recurrent model would have a similar amount of opaque FLOP, since a full-context-length forward pass features a lot of opaque computation, even though it's low serial depth. However, if a swarm of agents could communicate with each other via opaque media, this would significantly increase the amount of opaque FLOP in the system. (Note that the diagram assumes that the single deep recurrent model is deployed for the same amount of time as the swarm. The swarm would have somewhat higher depth if it's deployed for longer and a shorter-lived agent can hand over an opaque state to later models.)}
  \label{fig:5}
\end{figure}

Define an \textbf{opaque system} to be an interaction of 1 or more AI models, not mediated by NL-rooted tokens. Beyond opaque serial depth, another interesting quantity to track is the maximum FLOP of any opaque system; this quantity lets us articulate why a swarm of AI models that communicate using opaque state is more concerning than a single AI model that thinks in opaque state, as shown in Figure~\ref{fig:5}.

Ideally, AI developers would avoid creating models with high opaque serial depth. But, if that's intractable, we think it'd still be good to at least avoid building agent swarms that communicate using opaque state. However, we note that the switch to deep recurrent models may involve a lot of research and engineering work, whereas once you have deep recurrent models, it may be very easy to make them communicate with each other via opaque state.

Note that ``maximum FLOP of an opaque system'' scales linearly with individual models' FLOP, and is therefore being scaled much faster than opaque serial depth in the current CoT paradigm. So the main point of tracking this metric would be to capture the \emph{really large} jumps that may come from many different agents being wired together with opaque state, rather than more moderate scaling.

\section{Issues with low-FLOP serially intense computations}\label{appendix-g-issues-with-low-flop-serially-intense-computations}

\begin{figure}[t]
  \centering
  \includegraphics[width=0.9\linewidth]{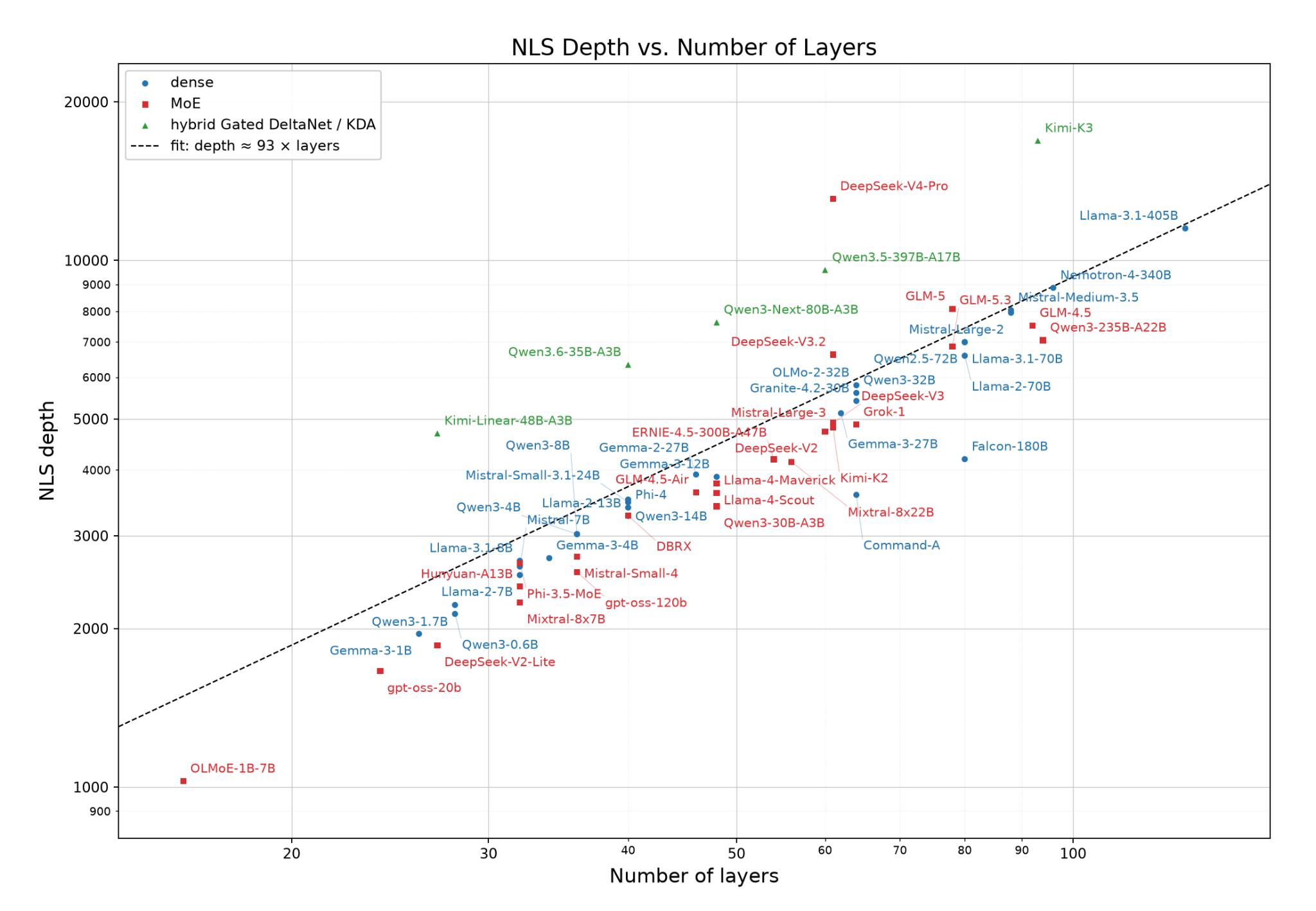}
  \caption[DeepSeek-V4-Pro has higher NLS depth than other similar architectures.]{DeepSeek-V4-Pro has higher NLS depth than other similar architectures.}
  \label{fig:6}
\end{figure}

Among models that don't use DeltaNet architectures, DeepSeek-V4-Pro is an outlier on NLS depth, as shown in Figure~\ref{fig:6}.

DeepSeek-V4-Pro's depth is large due to some very serially-intensive but low-FLOP computations that it performs. In particular, for every layer, it performs two extended iterated normalization operations (the \href{https://en.wikipedia.org/wiki/Sinkhorn\%27s_theorem\#Sinkhorn\%E2\%80\%93Knopp_algorithm}{Sinkhorn-Knopp algorithm}), which each contribute 100 to our serial depth. Across 61 layers, this corresponds to $100 \cdot 2 \cdot 61 = 12{,}200$ depth which is the dominant term in our computation of DeepSeek's NLS depth.

Importantly, each of these normalization operations take in only 16 FP32's produced by earlier layers and use no learned parameters, so it is clearly not capable of any dangerous opaque cognition. Ignoring the contributions of these normalization operations to the computational graph, the deepest remaining path is 25\% shorter.

We see the fact that DeepSeek-V4-Pro has high NLS depth as a flaw in our definition of NLS depth. However, we think that this does not invalidate the core motivation behind reporting some version of opaque serial depth as a proxy, for a few different reasons.

For DeepSeek-V4-Pro in particular, there's a clear argument for why the Sinkhorn-Knopp computations should be excluded for the purposes of estimating its opaque capabilities. It's not a case of ambiguity as much as apparent misclassification. If an AI company wants to report opaque serial depth using a notion like NLS depth, and doesn't want an algorithm like Sinkhorn-Knopp to be counted towards their depth in the public eye, they should have no problem justifying an analogous argument while disclosing minimal details (to the public or a 3rd party evaluators). Such as only facts about how many FLOP and learned parameters the serially intensive computations make use of.\footnote{Less importantly, we suspect that it would have little or no performance cost to avoid the serially intensive computation in DeepSeek-V4-Pro. For example, by computing the normalization to a lower degree of precision, or by replacing it with a different algorithm. We haven't reviewed it in depth, but we have noticed a paper addressed specifically to fix the DeepSeek architecture aptly titled ``\href{https://arxiv.org/abs/2601.05732v1}{mHC-lite: You Don't Need 20 Sinkhorn-Knopp Iterations}'' \citep{yang2026mhc}.}

We also believe this particular issue could have been addressed by reporting numbers from a variant of NLS depth with higher-bit gates in the circuit, which isn't entirely unprincipled, though it would require some other modifications to the definition.\footnote{If the circuits were constructed using arbitrary functions between 256-bit gates rather than 32-bit gates, all iterations in one instance of the Sinkhorn-Knopp algorithm could have fit into one such function, and the normal operations of the transformer would once again dominate the depth. However, increasing the bit rate would have some issues that would require some modifications to other parts of the definition. At the very least, parameters would have to be taken as inputs rather than be wired into the circuit, so that entire learned components of the neural network with a small number of outputs and inputs couldn't be fully compressed into a single step. (This could probably be specified by requiring a model's computational graph to be identical throughout training.) With these corresponding adjustments, a change in bit-rate would not be entirely unprincipled as a method for handling this kind of problem. In particular, it would allow for reducing serial depth most drastically when a serially deep algorithm has very few inputs and learned parameters, which also means that it wouldn't be able to substantially contribute to serially opaque cognition. However, it wouldn't be a complete fix. For example, a Sinkhorn-Knopp computation that normalized a very large table would require a yet-larger bit-rate, and it wouldn't be feasible to set a bit-rate so high that it addresses all problems like this.}

Overall, we don't think that the specific algorithm in DeepSeek-V4-Pro poses any substantial problems given realistic options for adjusting the metric or justifying exceptions on a case-by-case basis. But is there a risk that future edge cases will be more ambiguous, and harder to evaluate?

A weak reason to expect clear artifacts to be more common than actual edge cases is that AI developers are likely more comfortable with gratuitous looping for extremely cheap computations. There's more incentive to optimize and parallelize for more intensive ones.

However, that's not a rock-solid argument. And in fact, we do have an example of an architecture that's more of an edge case. Kimi-K3's NLS depth is larger because Kimi-K3 uses ``Kimi Delta Attention'', a form of state space model / linear attention to scale the depth above what you'd get in a normal transformer of its size. We think that Kimi-K3 is in fact somewhat concerning from a monitorability perspective compared to normal transformers in virtue of having larger NLS depth---but likely still not as concerning as it would be if it achieved a similar depth by just stacking layers.

Still, at minimum, the factors of overestimation involved (for both DeepSeek and the DeltaNets) remain small compared to the factors involved in the most concerning architecture shifts (which would remove the NL-rooted bottleneck at every autoregressive call). So tracking opaque serial depth will at least be useful as information about these most concerning changes.

And beyond this, it's plausibly tractable to develop a better notion of opaque serial depth that is more responsive to our actual concern. In particular, it may be possible to use a version that somehow quantitatively weighs different serial paths by their FLOP-count, and thereby directly captures the intuitive difference between different concerns. We're excited about further work on better proxies.

With the current state of information, we would still expect NLS depth (or similar notions) to be a significantly useful complement for understanding models' likely monitorability. Increased information sharing and research on the topic will hopefully improve our understanding about exactly how useful it is.

\end{document}